\documentclass[10pt,journal,compsoc]{IEEEtran}

\ifCLASSOPTIONcompsoc
  \usepackage[nocompress]{cite}
\else
  \usepackage{cite}
\fi

\ifCLASSINFOpdf
\else
\fi

\usepackage[caption=false,font=footnotesize,labelfont=rm,textfont=rm]{subfig}
\usepackage{textcomp}
\usepackage{stfloats}
\usepackage{url}
\usepackage{verbatim}
\usepackage{graphicx}
\usepackage{cite}
\usepackage{multirow}
\usepackage{amsmath,amsfonts}
\usepackage{algorithmic}
\usepackage{algorithm}
\usepackage{float}

\begin{document}
\title{Understanding the Effects of Projectors in Knowledge Distillation}

\author{Yudong Chen,
        Sen Wang,
        Jiajun Liu,
        Xuwei Xu,
        Frank de Hoog,
        Brano Kusy,
        and Zi Huang
\IEEEcompsocitemizethanks{\IEEEcompsocthanksitem Y. Chen and X. Xu are
with School of Electrical Engineering and Computer Science, The University of Queensland, Australia, and also with CSIRO Data61, Australia.\protect\\
E-mail: $\{$yudong.chen, xuwei.xu$\}$@uq.edu.au
\IEEEcompsocthanksitem S. Wang and Z. Huang are
with School of Electrical Engineering and Computer Science, The University of Queensland, Australia.\protect\\
E-mail: $\{$sen.wang, helen.huang$\}$@uq.edu.au
\IEEEcompsocthanksitem J. Liu, F. de Hoog and B. Kusy are with CSIRO Data61, Australia.\\
E-mail: $\{$jiajun.liu, frank.dehoog, brano.kusy$\}$@csiro.au}
\thanks{Manuscript received April 19, 2005.}}

\markboth{Journal of \LaTeX\ Class Files,~Vol.~14, No.~8, August~2015}%
{Shell \MakeLowercase{\textit{et al.}}: Bare Advanced Demo of IEEEtran.cls for IEEE Computer Society Journals}

\IEEEtitleabstractindextext{%
\begin{abstract}
Conventionally, during the knowledge distillation process (e.g. feature distillation), an additional projector is often required to perform feature transformation due to the dimension mismatch between the teacher and the student networks. Interestingly, we discovered that even if the student and the teacher have the same feature dimensions, adding a projector still helps to improve the distillation performance. In addition, projectors even improve logit distillation if we add them to the architecture too. Inspired by these surprising findings and the general lack of understanding of the projectors in the knowledge distillation process from existing literature, this paper investigates the implicit role that projectors play but so far have been overlooked. Our empirical study shows that the student with a projector (1) obtains a better trade-off between the training accuracy and the testing accuracy compared to the student without a projector when it has the same feature dimensions as the teacher, (2) better preserves its similarity to the teacher beyond shallow and numeric resemblance, from the view of Centered Kernel Alignment (CKA) \cite{CKA}, and (3) avoids being over-confident \cite{calibration} as the teacher does at the testing phase. Motivated by the positive effects of projectors, we propose a projector ensemble-based feature distillation method to further improve distillation performance. Despite the simplicity of the proposed strategy, empirical results from the evaluation of classification tasks on benchmark datasets demonstrate the superior classification performance of our method on a broad range of teacher-student pairs and verify from the aspects of CKA and model calibration that the student's features are of improved quality with the projector ensemble design. Code is available at \url{https://github.com/chenyd7/PEFD}.
\end{abstract}

}

\maketitle

\IEEEdisplaynontitleabstractindextext

\IEEEpeerreviewmaketitle

\ifCLASSOPTIONcompsoc
\IEEEraisesectionheading{\section{Introduction}\label{sec:introduction}}
\else
\section{Introduction}
\label{sec:introduction}
\fi

\IEEEPARstart{D}{eep} Neural Networks (DNNs) have become increasingly important in the field of computer vision because of their powerful feature extraction ability from images \cite{alexnet,inception,resnet,convnext}. The commonly used DNNs usually require high computation complexity and contain a large number of parameters. Therefore, it is inefficient to deploy these high-capacity networks on edge devices, such as mobile phones and embedded equipment, which have limited computing power and memory resources. To alleviate this problem, compact neural networks with fewer parameters have been developed, such as MobileNet \cite{mobilenets} and ShuffleNet \cite{shufflenet}. However, there is inevitably a performance gap between high-capacity networks and compact networks. To narrow this gap, Knowledge Distillation (KD) \cite{kd,pami1,pami2,pami4} is utilized. KD methods aim to introduce the hidden information in the high-capacity network (teacher) to assist the training of the compact network (student) and improve its performance. KD is verified to be effective across a wide range of different DNN architectures that include Convolutional NNs \cite{ofd,pefd}, and Transformer-based NNs \cite{deit,spkd5}. 

Existing distillation methods can be roughly categorized into logit-based, feature-based, and similarity-based distillation \cite{kdsurvey}. Recent research shows that when used alone, feature-based methods generally distill a better student network compared to the other two groups \cite{crd,cid,srrl,srrl2}. To improve the generalizability of the student, various feature distillation methods have been developed by designing more powerful objective functions \cite{crd,pad,cid,srrl,srrl2,mgd} and determining more effective correspondences between the layers of the student and the teacher \cite{kr,afd,semckd}. 
However, we identify a large research gap in existing literature, that is the underlying role and mechanism that the (feature) projectors have in the KD process.

\begin{figure}[t]
\centering
\subfloat[w/o Projector]{\includegraphics[scale=0.15,trim=8 7 7 8, clip]{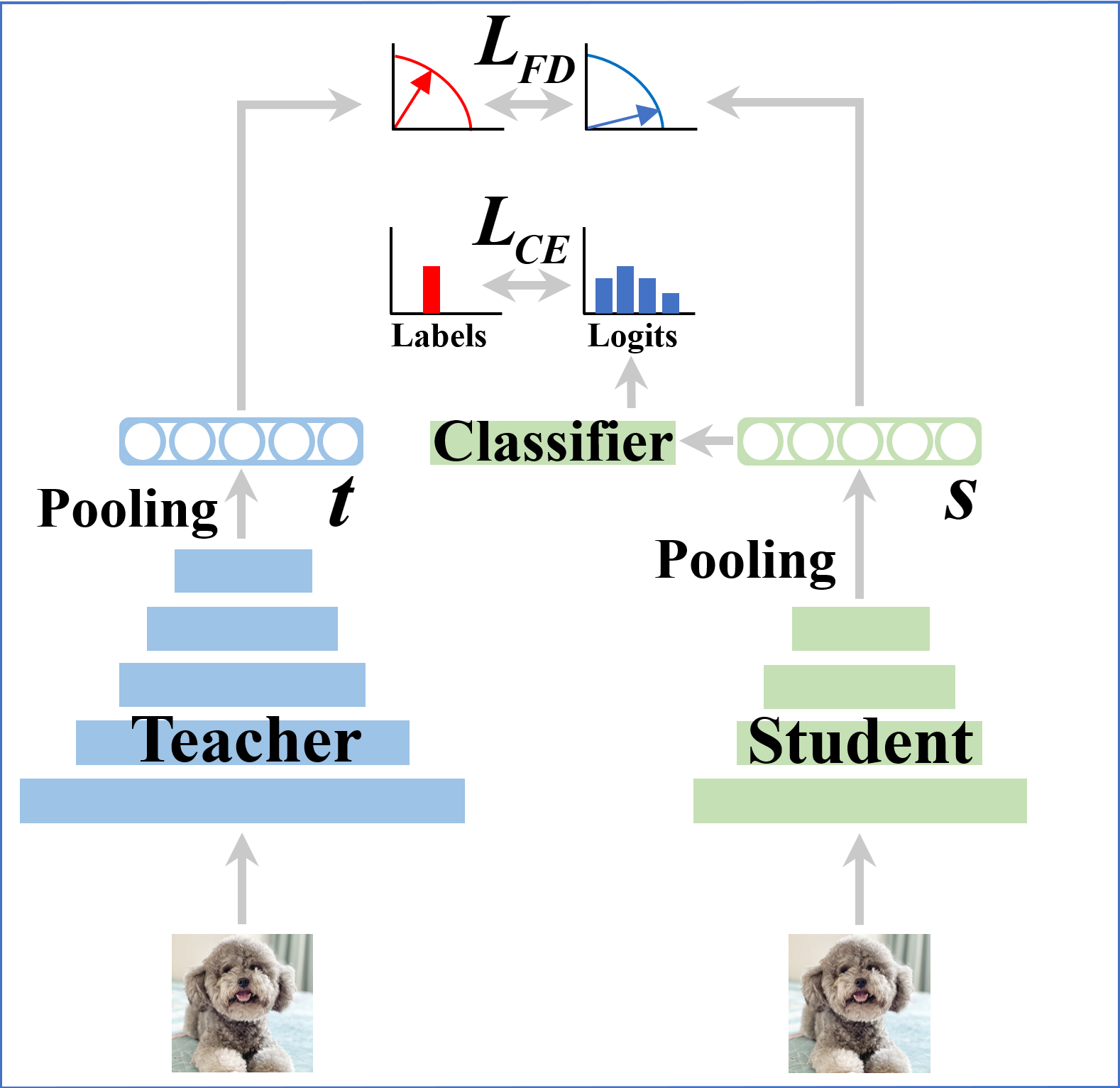}%
\label{pipeline0}}
\hspace{3mm}
\subfloat[w/ Projector]{\includegraphics[scale=0.15,trim=8 7 7 8, clip]{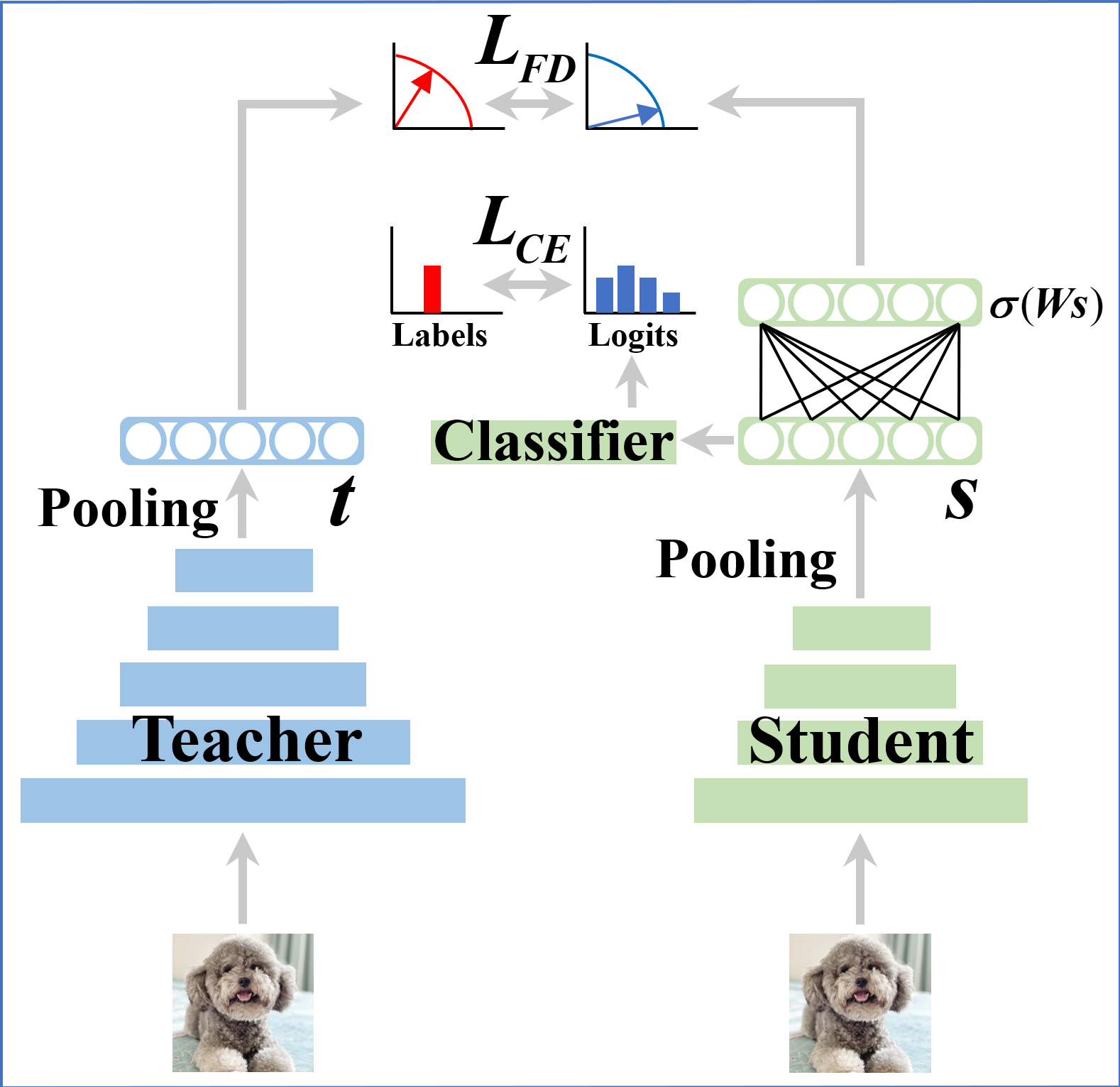}%
\label{pipeline1}}
\caption{Illustration of (a) feature distillation without a projector, and (b) the general feature-based distillation with a projector \cite{cid,srrl,srrl2}, where $s$, $t$, $\sigma(Ws)$, $W$, $\sigma(\cdot)$, $\mathcal{L}_{CE}$ and $\mathcal{L}_{FD}$ are the student feature, the teacher feature, the transformed student feature, the weighting matrix, the ReLU function, the cross-entropy loss, and the feature distillation loss, respectively.}
\label{pipeline}
\end{figure}

More often than not, the feature dimensions of the student networks do not agree with that of the teacher networks, and as a result, projectors are required to perform feature transformations to align feature dimensions to enable distillation. So far this has been the only and limited understanding of projectors in KD. However, we have found that even if the student and the teacher have the same feature dimensions, imposing a projector on the student network can still improve the distillation performance as shown in Figure \ref{tradeoff} and in \cite{pefd}. We conjecture that the use of projectors helps to relax the student from strictly numeric alignment to the teacher's outputs, and allows the student to focus more on learning from the teacher's essential knowledge about the task. As shown in Figure \ref{pipeline0}, distillation without a projector can be regarded as a multi-task learning process in a shared student's feature space, including feature learning for classification and feature alignment for distillation from the same sub-network. In this case, if the student pays more attention to the classification task, the teacher's supervision will be weakened and the generalizability of the student cannot be effectively improved. On the contrary, if we assign a larger weight to the distillation task, the student tends to underfit the ground-truth classification labels. 
The underfitting problem occurs because the student is set to be optimized directly to exactly mimic the teacher's feature distribution and the student does not have enough capacity to achieve this goal. As such, the sub-optimal solutions cannot satisfy the minimization of the distillation and classification losses at the same time \cite{kd1}. Since the classification and the distillation tasks are coupled in the same sub-network, and the distillation task is assigned a larger weight, this limits the student's ability to learn from the classification task.
By adding a projector into the student network as illustrated in Figure \ref{pipeline1}, we can disentangle the two learning tasks and the main student sub-network before the projector does not need to mimic the teacher's feature distribution directly. With the update of the projector, the teacher's information can be adaptively modified and the student can receive more appropriate knowledge that is beneficial to the optimization of the classification task, while still retaining extracted knowledge from the teacher beyond numeric values. 

Another interesting finding is that the decoupling effects from projectors also exist for the original logit-based KD. Specifically, instead of feeding the student's logits into the distillation loss and the cross-entropy loss simultaneously, we utilize a projector to transform the logits generated by the student and feed the transformed logits to the distillation loss. This consistently improves KD's performance as well. We interpret this phenomenon from the perspective of a better trade-off between the teacher's target class knowledge and non-target class knowledge \cite{kd6}. 

As a deep investigation to reveal the project's role and effects in the KD process, this paper presents a series of novel empirical studies. Followed by that, also motivated by the above findings, we further propose a better design of projectors to improve KD and offer a comprehensive quantitative evaluation for it. In summary, the contributions of this paper are as follows: 
\begin{enumerate}
    \item \textbf{We conduct an in-depth empirical analysis of the underlying mechanism of projectors in knowledge distillation.} We conjecture that projectors help to extract more appropriate knowledge from the teacher for the training of the student. This phenomenon is observed from three aspects: (1.1) \textbf{Firstly, the use of projectors enables the student to focus more on the teacher's knowledge which is beneficial to the main classification task.} We show that the joint optimization of distillation and classification in one student space limits the effectiveness of the classification loss. By adding a projector to decouple the two losses, the projector takes more responsibility for aligning the student's outputs to the teacher's outputs numerically, while the student sub-network can now focus more on learning appropriate knowledge that improves classification. We verify this claim based on the comparisons of training and testing accuracies of the student distilled with and without a projector by using feature and logit distillation methods. In addition, we demonstrate that the use of projectors makes the student pay more attention to the teacher's non-target class knowledge, which is regarded as important information for the success of KD \cite{kd6}. (1.2)\textbf{ Secondly, projectors facilitate the similarity preservation of the teacher's intrinsic feature distribution.} Beyond shallow numeric values matching between student and teacher features, we investigate the Centered Kernel Alignment (CKA) \cite{CKA} similarities between the student and the teacher. From the measurements of CKA similarities, it is shown that the student better learns the teacher's intrinsic feature distribution when training with an additional projector. (1.3) \textbf{Thirdly, projectors mitigate the teacher's over-confidence at the testing phase.} We observe that the student distilled with a projector is better calibrated in terms of the consistency between confidence scores and accuracy of its predictions on testing sets. On the other hand, since the student distilled without a projector directly mimics the teacher's outputs in the original feature space, its model calibration property is affected by the over-confident teacher. 
    \item \textbf{We propose an ensemble-based projector design to improve the feature distillation and present a thorough evaluation.} Based on the observations of the benefits of adding a projector, we propose to ensemble multiple projectors to extract the teacher's information from different views and further improve the classification performance, similarity preservation, and model calibration. With this simple yet effective modification, the proposed projector ensemble consistently outperforms state-of-the-art distillation methods on different datasets with various teacher-student pairs.
\end{enumerate}

Part of the results were published previously in our conference paper \cite{pefd} but this paper presents a wide range of significant and novel findings as follows: (1) By redesigning the experiments, more convincing results are added to analyze the effects of projectors in feature distillation and an improved logit distillation method is proposed based on these observations. (2) To monitor the feature similarities between the student and the teacher, we compute the CKA similarities during training, which provides a new angle to study the effects of projectors and complements the conference paper. (3) We additionally examine the model calibration problem for the distillation with and without a projector. (4) More experiments are conducted to investigate the effectiveness of the proposed method with different initialization strategies and compare the transferability of different distilled students on benchmark datasets. In addition, ablation studies on ImageNet \cite{imagenet} are also provided. 

The remaining sections are organized as follows. In Section \ref{relatedwork}, we review some related distillation methods. Section \ref{methodsection} discusses the effects of projectors and introduces the proposed method. Experimental results are given in Section \ref{experimentsection}. Section \ref{conclusection} summarizes the whole manuscript.

\section{Related Work} \label{relatedwork}
Since this paper mainly focuses on the mechanism of projectors, we divide the existing methods into two categories in terms of the usage of projectors, i.e., projector-free distillation and projector-dependent distillation.
\subsection{Projector-free Distillation}
As the most representative distillation method, the original KD \cite{kd} utilizes the logits generated by the pre-trained teacher to be the additional targets of the student. The intuition of KD is that the generated logits provide more useful information than the general binary labels (ground-truth labels) for optimization. Motivated by the success of KD, various logit-based methods have been proposed for further improvement \cite{dml,takd,pami5,tip1,tip2,kd1,kd2,kd3,kd4,kd5,kd6,kd7,kd8}. For example, Deep Mutual Learning (DML) \cite{dml} proposes replacing the pre-trained teacher with an ensemble of students so that the distillation mechanism does not need to train a large network in advance. Teacher Assistant Knowledge Distillation (TAKD) \cite{takd} observes that a better teacher may distill a worse student due to the large performance gap between them. Therefore, a teacher assistant network is introduced to alleviate this problem. Another technical route of projector-free methods is similarity-based distillation \cite{sp,cc,spkd1,spkd2,spkd3,spkd6,spkd4,spkd5}. Unlike the logit-based methods that aim to exploit the category information hidden in the predictions of the teacher, similarity-based methods explore the latent relationships between samples in feature space. For example, Similarity-Preserving (SP) \cite{sp} distillation first constructs the similarity matrices of the student and the teacher by computing the inner products between features and then minimizes the discrepancy between these similarity matrices. Correlation Congruence (CC) \cite{cc} forms the similarity matrices with a kernel function. Although the logit-based and similarity-based methods do not require an extra projector during training, they are generally less effective than the feature-based methods as shown in the recent research \cite{cid,srrl2}.  

\subsection{Projector-dependent Distillation} The original feature distillation method FitNets \cite{fitnets} minimizes the L2 distances between student and teacher feature maps produced by the intermediate layer of networks. Furthermore, Contrastive Representation Distillation (CRD) \cite{crd}, Softmax Regression Representation Learning (SRRL) \cite{srrl2,srrl}, and Comprehensive Interventional Distillation (CID) \cite{cid} show that the last feature representations of networks are more suitable for distillation. One potential reason is that the last feature representations are closer to the classifier and will directly affect the classification performance \cite{srrl}. The aforementioned feature distillation methods mainly focus on the design of loss functions such as introducing contrastive learning \cite{crd} and imposing causal intervention \cite{cid}. A simple 1x1 convolutional kernel or a linear projection is adopted to transform features in these methods. We note that the effect of projectors is under-explored. Previous works such as Factor Transfer (FT) \cite{ft} and Overhaul of Feature Distillation (OFD) \cite{ofd} try to improve the architecture of projectors by introducing an auto-encoder and modifying the activation function. However, their performance is not competitive when compared to the state-of-the-art methods \cite{srrl2,cid}. Instead, this paper proposes a simple distillation framework by combining the ideas of distilling the last features and projector ensemble.

\begin{figure}[t]
\centering
\subfloat[ResNet20x4-ResNet8x4]{\includegraphics[scale=0.31]{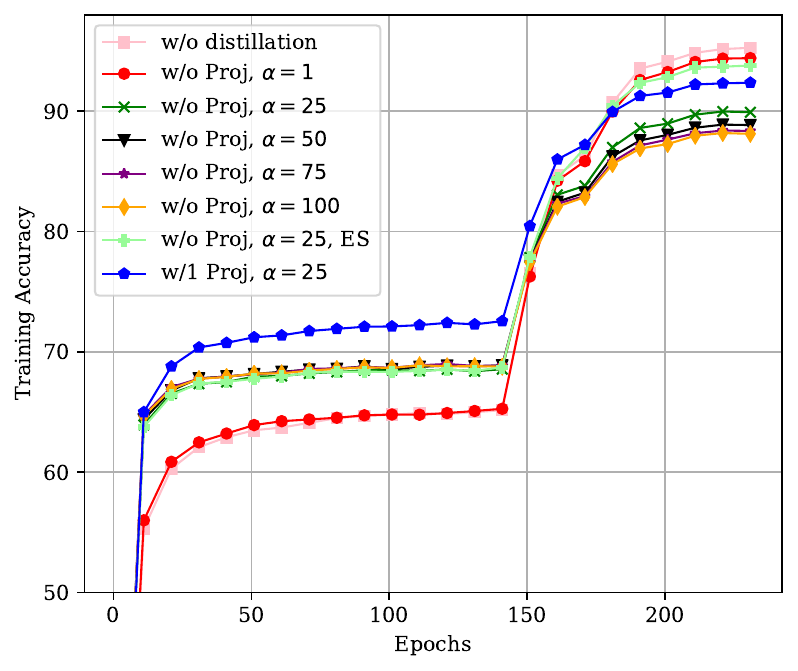}%
\label{to0}}
\hspace{2mm}
\subfloat[ResNet20x4-ResNet8x4]{\includegraphics[scale=0.31]{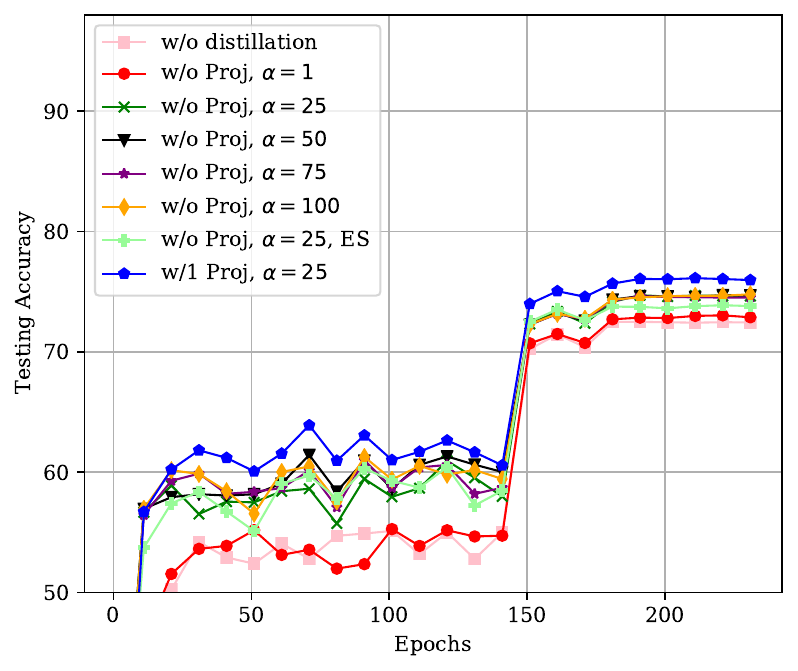}%
\label{to1}}
\\
\subfloat[ResNet32x4-ResNet8x4]{\includegraphics[scale=0.31]{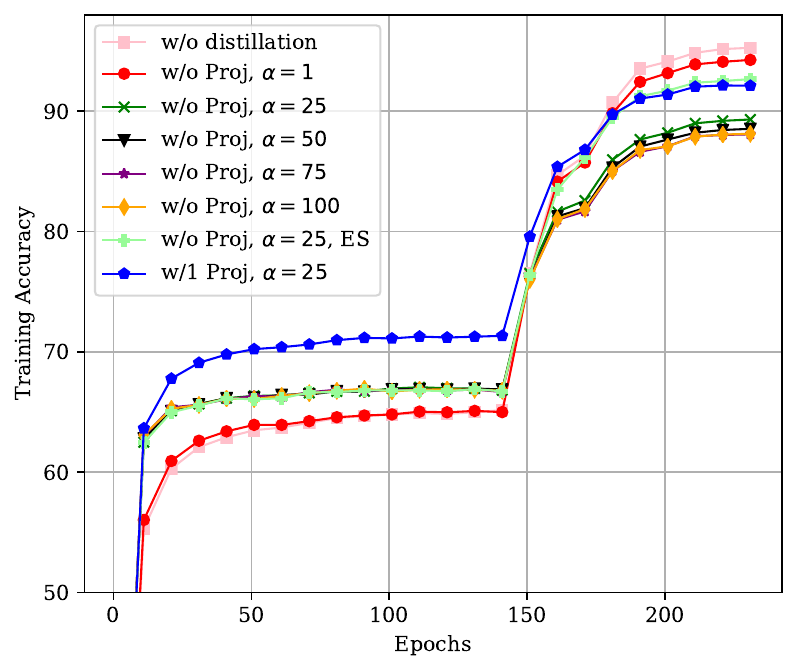}%
\label{to2}}
\hspace{2mm}
\subfloat[ResNet32x4-ResNet8x4]{\includegraphics[scale=0.31]{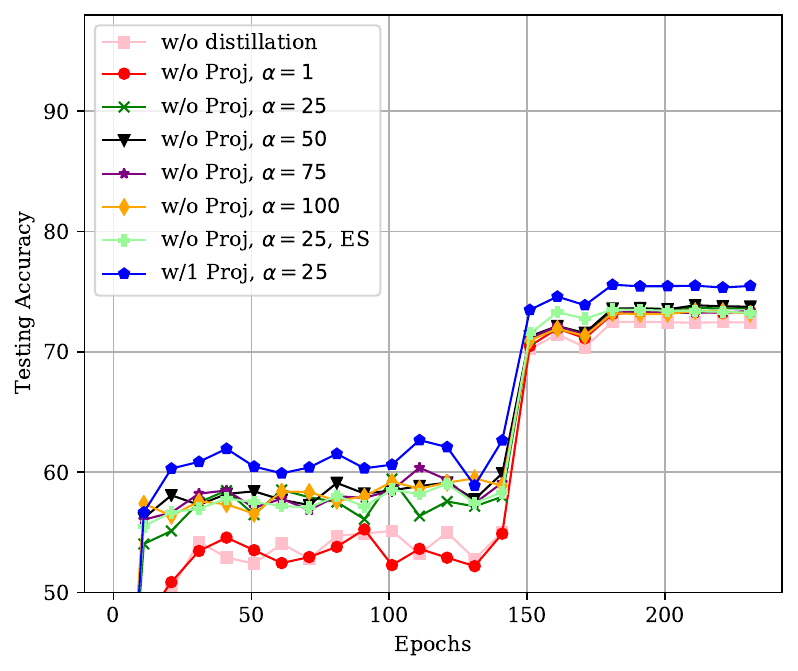}%
\label{to3}}
\caption{Training and testing accuracy curves by using feature distillation on CIFAR-100 dataset with different teacher-student pairs. In this experiment, the feature dimensions of the student and the teacher are the same, which means $d=m$.}
\label{tradeoff}
\end{figure}

\section{Underlying Effects of Projectors}
\label{methodsection}
We first define the notations used in the following sections. In line with observations in recent research \cite{crd,cid}, we apply the feature distillation loss to the layer before the classifier. $S=\{ s_{1},s_{2},...,s_{i},...,s_{b}\} \in \mathbb{R}^{d\times b}$ denotes the last student features, where $d$ and $b$ are the feature dimension and the batch size, respectively. The corresponding teacher features are represented by $T=\{ t_{1},t_{2},...,t_{i},...,t_{b}\} \in \mathbb{R}^{m\times b}$, where $m$ is the feature dimension. To match the dimensions of $S$ and $T$, a projector $g(\cdot)$ is required to transform the student or teacher features. We have observed that imposing the projector on the teacher network is less effective. Therefore, in the proposed distillation framework, a projector will be added to the student as $g(s_i)=\sigma(Ws_i)$ during training and be removed after training, where $\sigma(\cdot)$ is the ReLU function and $W \in \mathbb{R}^{m\times d}$ is a weighting matrix.

\subsection{Preliminaries}
The feature distillation loss proposed in FitNets is to minimize the L2 distances between student and teacher features. However, such an objective is sensitive to the length of the feature vector \cite{cid}. To alleviate this problem, we normalize feature vectors in advance and adopt the following Direction Alignment (DA) loss \cite{at,da1,da2} for feature distillation:
\begin{equation}
\begin{split}
\mathcal{L}_{DA} &= \frac{1}{2b}\sum_{i=1}^b||\frac{g(s_i)}{||g(s_i)||_2}-\frac{t_i}{||t_i||_2}||_2^2
\\
&=1- \frac{1}{b}\sum_{i=1}^b\frac{\langle g(s_i),t_i\rangle}{||g(s_i)||_2||t_i||_2},
\end{split}
\label{fd}
\end{equation}
where $||\cdot||_2$ is the L2-norm and $\langle \cdot,\cdot\rangle$ represents the inner product of two vectors. By convention \cite{kd,crd,cid}, the distillation loss is coupled with the cross-entropy loss to train a student as follows: 
\begin{equation}
    \mathcal{L}_{total} = \mathcal{L}_{CE} + \alpha \mathcal{L}_{DA},
    \label{totaloss}
\end{equation}
where $\mathcal{L}_{CE}$ is the cross-entropy loss and $\alpha$ is a hyper-parameter used to control the weight of the DA loss.

\subsection{Projectors Improve Feature Distillation}
As mentioned in the Introduction, the student benefits from adding a projector when the feature dimensions of the student and the teacher are the same. We claim that the projector can be used to disentangle the classification and distillation tasks and extract more appropriate knowledge being passed to the student. To evaluate this claim, we start by jointly optimizing classification and distillation. To be specific, we remove the projector in Equation \ref{totaloss} and train two teacher-student pairs with different values of $\alpha$. 

Experimental results are shown in Figure \ref{tradeoff}. Our observations are as follows: (1) The classification loss and the distillation loss are both important to the student's overall performance. When we set $\alpha=0$, the student obtains the lowest testing accuracy on both teacher-student pairs, which indicates the effectiveness of feature distillation. On the other hand, when we gradually increase the value of $\alpha$, the student's testing accuracy gradually increases and tends to saturate in $\alpha=25$. If we further increase the value of $\alpha$, the student's performance tends to decrease. That is to say, if we mainly focus on the optimization of the distillation loss, the student is unable to achieve the best performance. These results demonstrate the importance of classification loss. (2) The effectiveness of the classification loss is limited under the joint optimization setting. According to the results on both teacher-student pairs, we can find that the training accuracy of the student constantly decreases with the increase of $\alpha$. This is because the student does not have enough capacity to simultaneously optimize the distillation and the classification losses and pays more attention to the optimization of the distillation loss \cite{kd1}. By solely optimizing the classification loss, the student obtains the highest training accuracy, which indicates the effectiveness of the classification loss is limited and can be further improved. (3) Early stopping (ES) of the distillation task degrades the student's test performance. As indicated in \cite{kd1}, due to the capacity mismatch between the student and the teacher, it is beneficial to improve the student's feature learning ability for classification by stopping the teacher's supervision at the early training stage. To verify this claim, we set $\alpha=25$ and stop distillation in the middle of training. From the results in Figure \ref{tradeoff}, we observe that discarding the teacher's supervision does increase the final training accuracy but decreases the testing accuracy at the same time. (4) Introducing a projector to the student helps to obtain a better trade-off between training accuracy and testing accuracy. As discussed in \cite{kd1}, the optimization dilemma between distillation and classification is because the student does not have enough capacity to fully mimic the teacher's feature distribution. In addition, since the classification and distillation losses are performed in a shared student feature space and the distillation loss is assigned a larger weight to obtain better performance, the effectiveness of the classification loss is suppressed. By adding a projector to transform student features and perform distillation in a different space, the student is relaxed from strictly mimicking the numeric values of the teacher's features and focuses more on the teacher's essential knowledge for classification. To validate this conjecture, we set $\alpha=25$ and train the student with a projector using Equation \ref{totaloss}. Figure \ref{tradeoff} illustrates that the student with a projector obtains higher training accuracy compared to the corresponding student without a projector. In addition, the testing accuracy of the student is also greatly increased. These results demonstrate that more appropriate teacher knowledge that is beneficial to the main classification task is transferred to the student by adding a projector.

\begin{figure}[!t]
\centering
\subfloat[w/o Projector]{\includegraphics[scale=0.168,trim=8 7 7 8, clip]{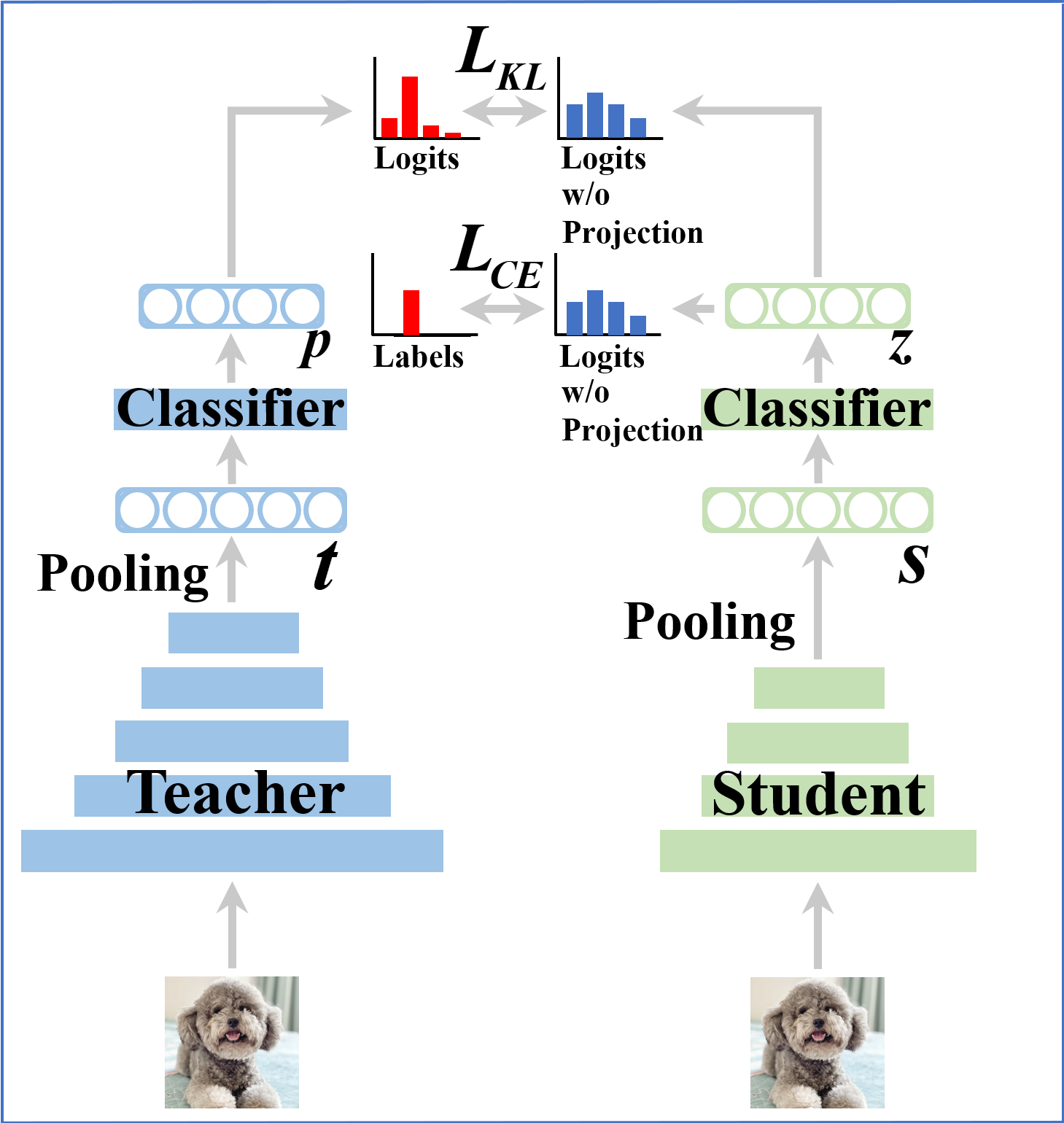}%
\label{pipelinekd0}}
\hspace{2mm}
\subfloat[w/ Projector]{\includegraphics[scale=0.168,trim=8 7 7 8, clip]{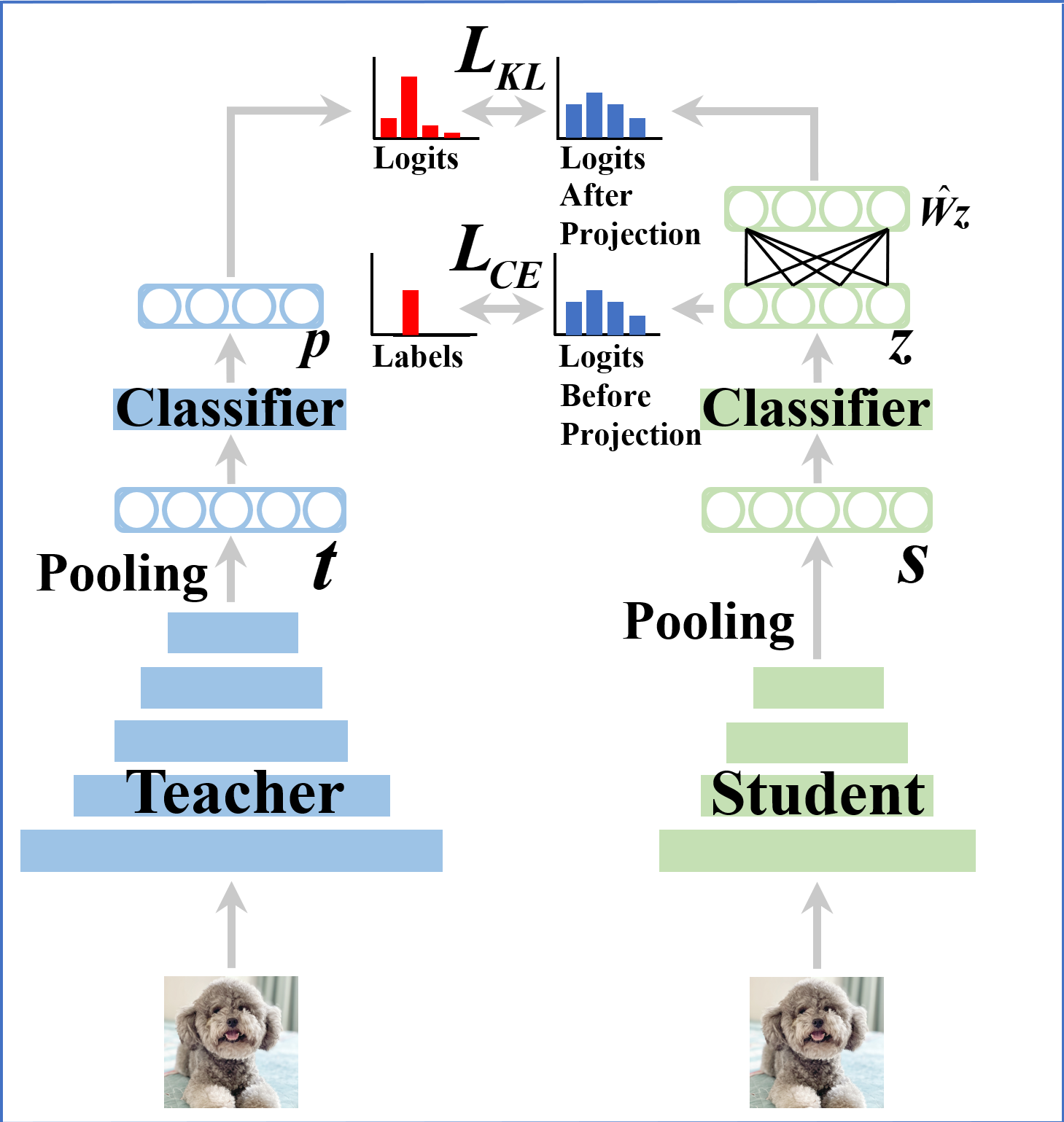}%
\label{pipelinekd1}}
\caption{Illustration of (a) logit distillation without a projector, and (b) with a projector, where $p$, $z$, $\hat{W}z$, $\hat{W}$ and $\mathcal{L}_{KL}$ are the teacher logit, the student logit, the transformed student logit, the weighting matrix, and the KL divergence loss, respectively.}
\label{pipelinekd}
\end{figure}

\subsection{Projectors Improve Logit Distillation}
The previous subsection discusses the potential mechanism of the projector in feature distillation and the benefits of applying a projector in the training process. To demonstrate the generality of our findings, we further introduce the projector to the original KD. As a projector-free method, KD does not need to add a projector to the student during training since the dimensions of student and teacher logits are always the same (i.e. the number of classes). The goal of KD is to let the student mimic the logits of the teacher and the objective function is as follows:
\begin{equation}
    \mathcal{L}_{KD} = \beta \mathcal{L}_{CE} + (1-\beta) \mathcal{L}_{KL},
    \label{totalossKD}
\end{equation}
where $\beta$ is a balance coefficient, and $\mathcal{L}_{KL}$ is the KL divergence between student and teacher logits, which is written as follows:
\begin{equation}
    \mathcal{L}_{KL} = -\sum_{i=1}^c\frac{\exp(p^i/\mu)}{\sum_j\exp(p^j/\mu)}\log \frac{\exp(z^i/\mu)}{\sum_j\exp(z^j/\mu)},
    \label{KLloss}
\end{equation}
where $c$ is the number of classes, $\mu$ is a temperature coefficient, $p^i$ is the $i$-th element of $p$, $z^i$ is the $i$-th element of $z$, and $p\in \mathbb{R}^{c\times 1}$ and $z\in \mathbb{R}^{c\times 1}$ are the teacher logit and the student logit of a sample, respectively. By adding a projector to transform the student logit, we have $v=\hat{W}z$, where $\hat{W} \in \mathbb{R}^{c\times c}$ is a weighting matrix. The modified KL loss is described as follows:
\begin{equation}
    \mathcal{L}_{MKL} = -\sum_{i=1}^c\frac{\exp(p^i/\mu)}{\sum_j\exp(p^j/\mu)}\log \frac{\exp(v^i/\mu)}{\sum_j\exp(v^j/\mu)}.
    \label{MKLloss}
\end{equation}

\begin{figure}[t]
\centering
\subfloat[ResNet20x4-ResNet8x4]{\includegraphics[scale=0.31]{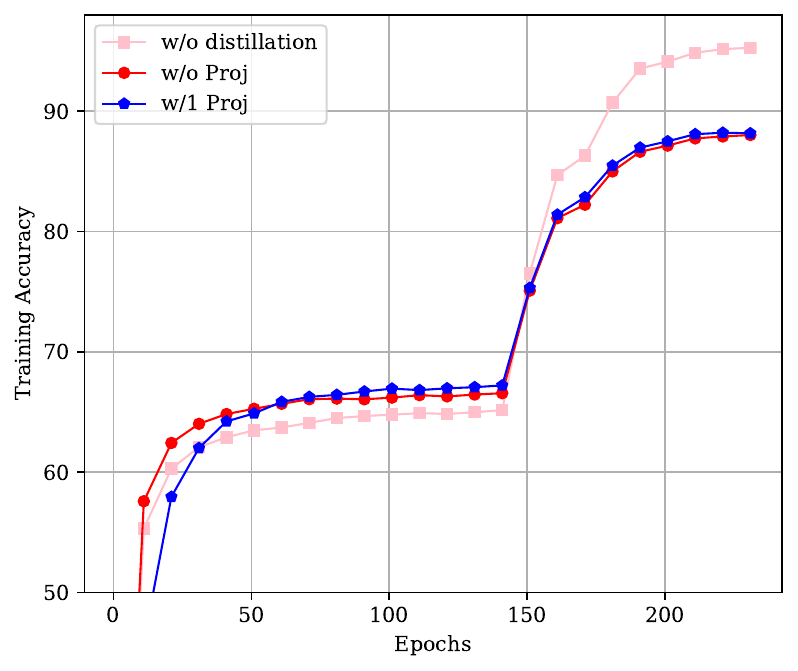}%
\label{kdproj0}}
\hspace{3mm}
\subfloat[ResNet20x4-ResNet8x4]{\includegraphics[scale=0.31]{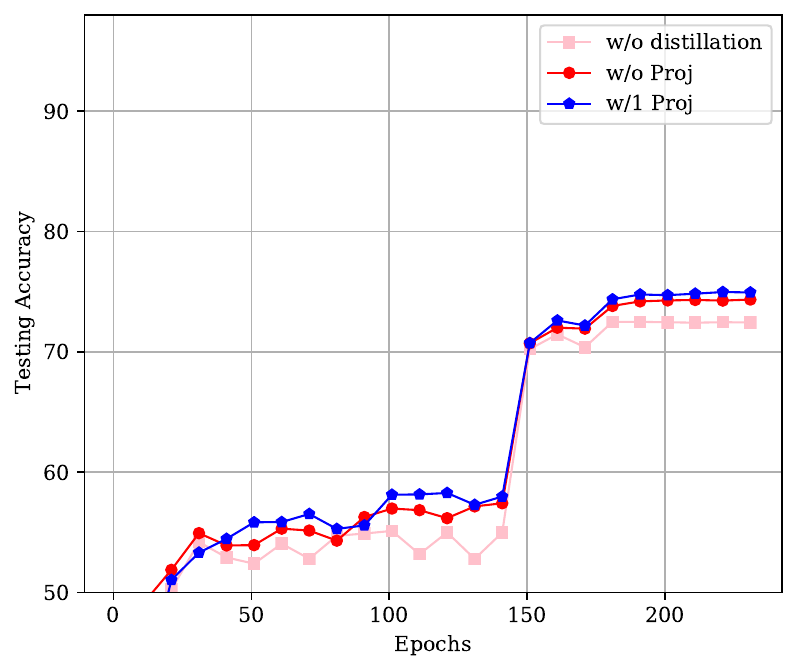}%
\label{kdproj1}}
\\
\subfloat[ResNet32x4-ResNet8x4]{\includegraphics[scale=0.31]{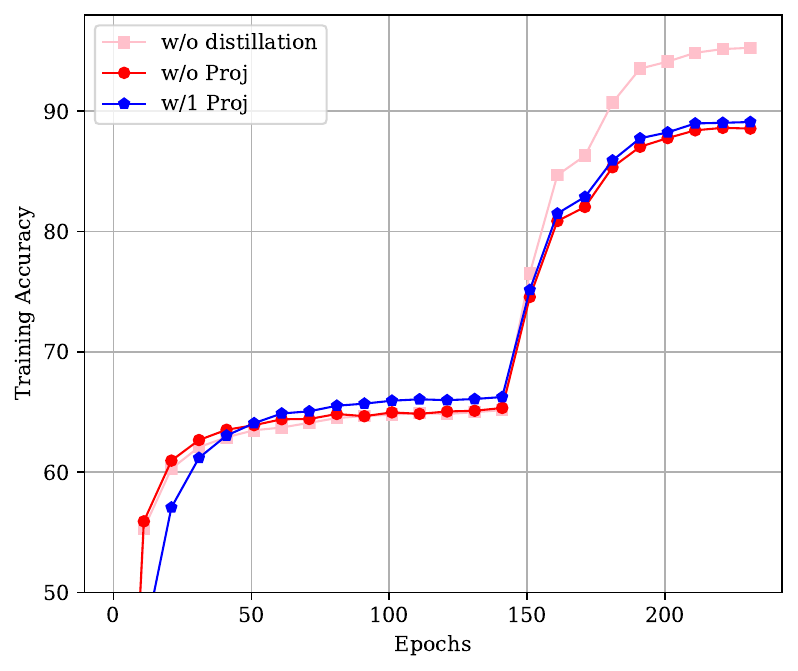}%
\label{kdproj2}}
\hspace{3mm}
\subfloat[ResNet32x4-ResNet8x4]{\includegraphics[scale=0.31]{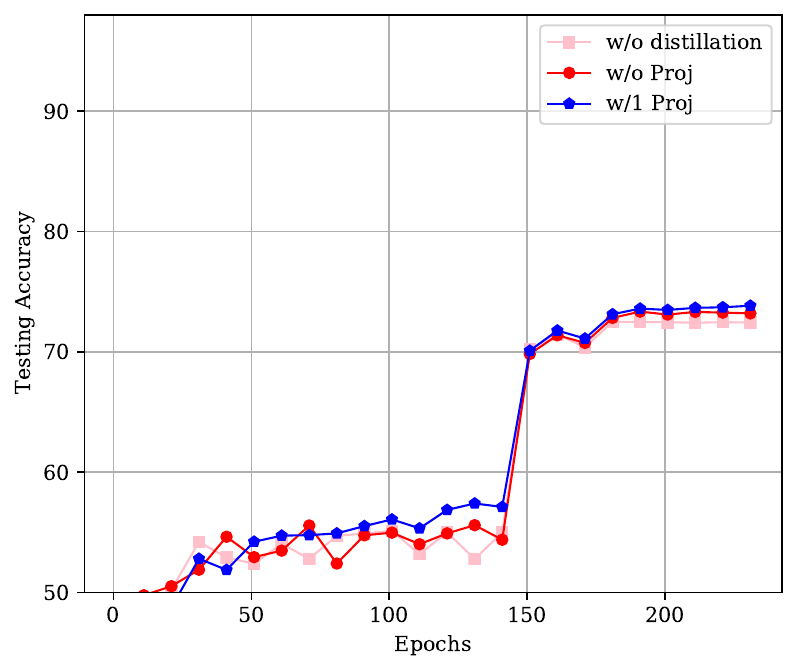}%
\label{kdproj3}}
\caption{Training and testing accuracy curves by using logit distillation on CIFAR-100 dataset with different teacher-student pairs.}
\label{kdproj}
\end{figure}

\begin{table}[t]
    \caption{TCKD and NCKD losses on CIFAR-100 training set by using logit distillation with ResNet32x4-ResNet8x4.}
    \centering
    \begin{tabular}{c|c|c|c} 
		\hline
            &\multirow{2}{*}{w/o Projector} &\multicolumn{2}{c}{w/ Projector}
            \\
            \cline{3-4}
		  & &Before Projection &After Projection\\
		\hline 
		TCKD 	&0.971	&4.341	&0.825\\
		NCKD 	&1.093	&1.014 &1.005\\
		\hline
    \end{tabular}
    \label{nckd}
\end{table}
The difference between the original and the modified logit distillation processes is illustrated in Figure \ref{pipelinekd}. Similar to the experiments on feature distillation, we use two teacher-student pairs to validate logit distillation with and without a projector. From the results in Figure \ref{kdproj}, we observe a similar phenomenon when using a projector to assist training. That is, the original logit distillation degrades the training accuracy of the student and the addition of a projector helps to improve the training accuracy as well as the testing accuracy of the student. 

In addition, we link our conjecture on the mechanism of projectors to the Target Class Knowledge Distillation (TCKD) and Non-target Class Knowledge Distillation (NCKD) losses proposed in Decoupled Knowledge Distillation (DKD) \cite{kd6}. As illustrated in \cite{kd6}, the original KD loss can be divided into TCKD and NCKD losses, and the authors observe that the strong coupling between them in the original KD loss tends to limit the optimization of NCKD loss and degrade the overall distillation performance. By introducing different hyper-parameters, DKD decouples the two losses and increases the effect of the NCKD loss. In this way, the logit distillation performance can be improved. In Table \ref{nckd}, we compute the NCKD and TCKD losses between the teacher and the student distilled with and without a projector. According to the results, we observe that the student without a projector tends to pay more attention to the TCKD loss which affects the optimization of the NCKD loss. These results are consistent with the observation in DKD. By adding a projector, the student mitigates this problem since the student logits before projection do not directly mimic the teacher logits and the student has more flexibility to select the teacher's information through the update of the projector. As such, the student can focus more on the optimization of the NCKD loss and become more effective in classification. This phenomenon further supports our claim that projectors extract more useful teacher knowledge for the student's training.

\begin{figure}[t]
\centering
\subfloat[ResNet20x4-ResNet8x4]{\includegraphics[scale=0.31]{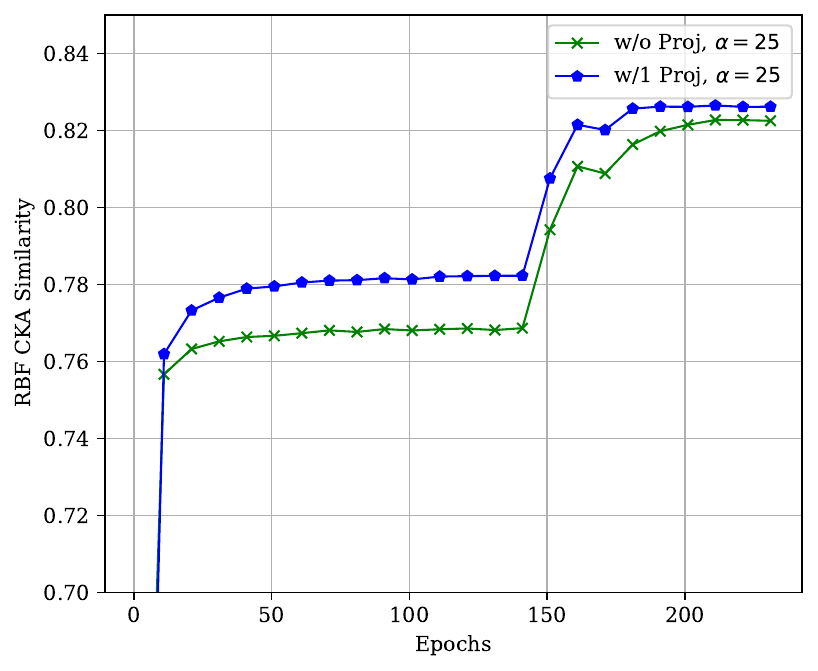}%
\label{cka1}}
\hspace{2mm}
\subfloat[ResNet32x4-ResNet8x4]{\includegraphics[scale=0.31]{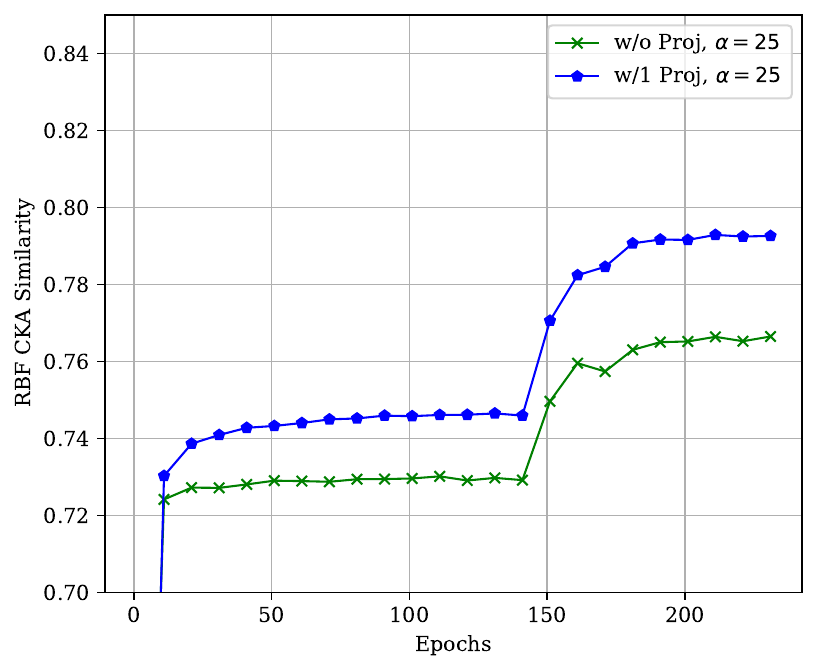}%
\label{cka3}}
\caption{CKA similarities between student features $S$ and teacher features $T$ on CIFAR-100 training set by distillation with and without a projector.}
\label{ckasimilarity}
\end{figure}

\begin{figure*}[t]
\centering
\subfloat[ResNet20x4]{\includegraphics[scale=0.28, trim=8 10 7 8, clip]{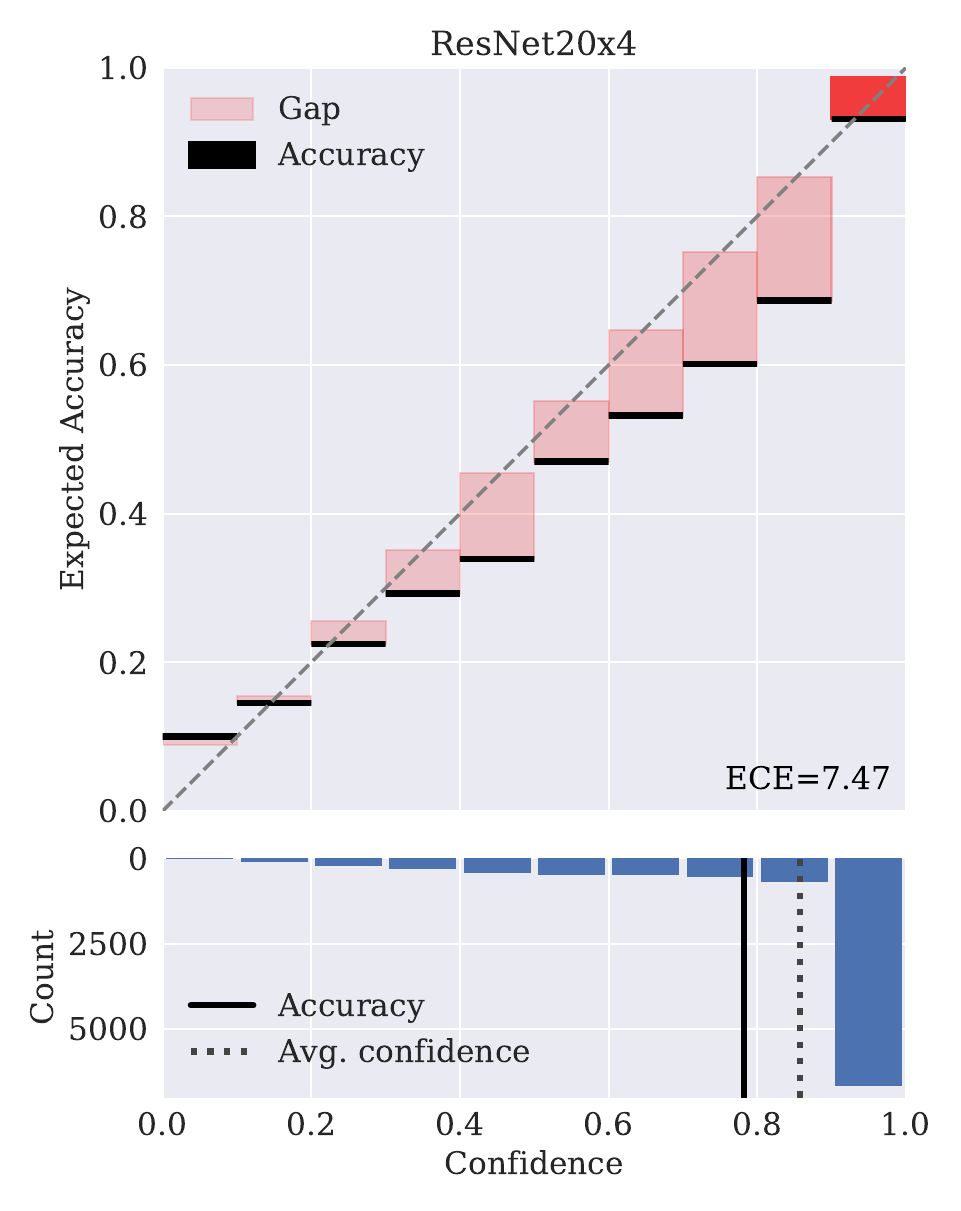}%
\label{ca0}}
\hspace{1mm}
\subfloat[ResNet8x4]{\includegraphics[scale=0.28, trim=8 10 7 8, clip]{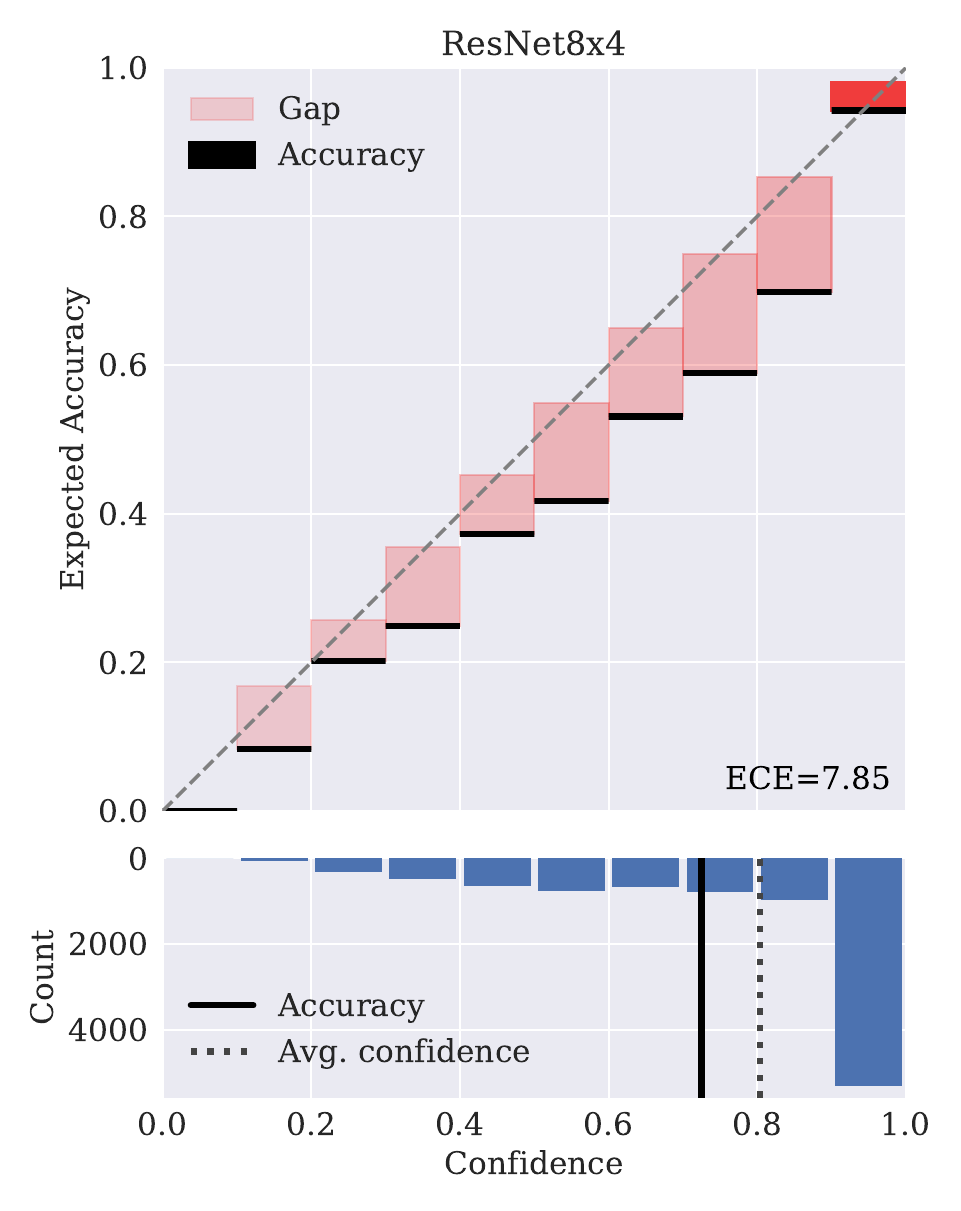}%
\label{ca1}}
\hspace{1mm}
\subfloat[w/o Projector]{\includegraphics[scale=0.28, trim=8 10 7 8, clip]{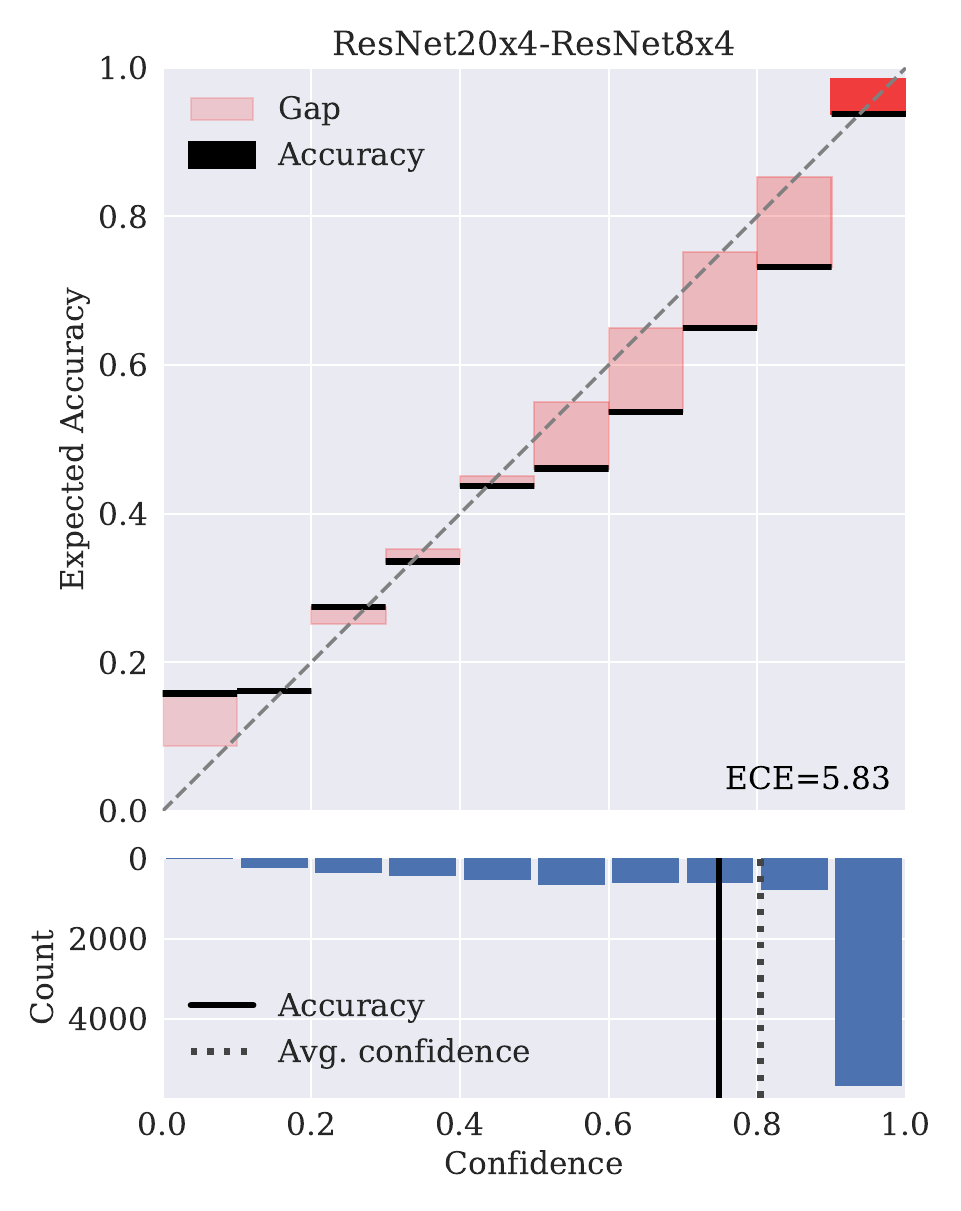}%
\label{ca2}}
\hspace{1mm}
\subfloat[w/1 Projector]{\includegraphics[scale=0.28, trim=8 10 7 8, clip]{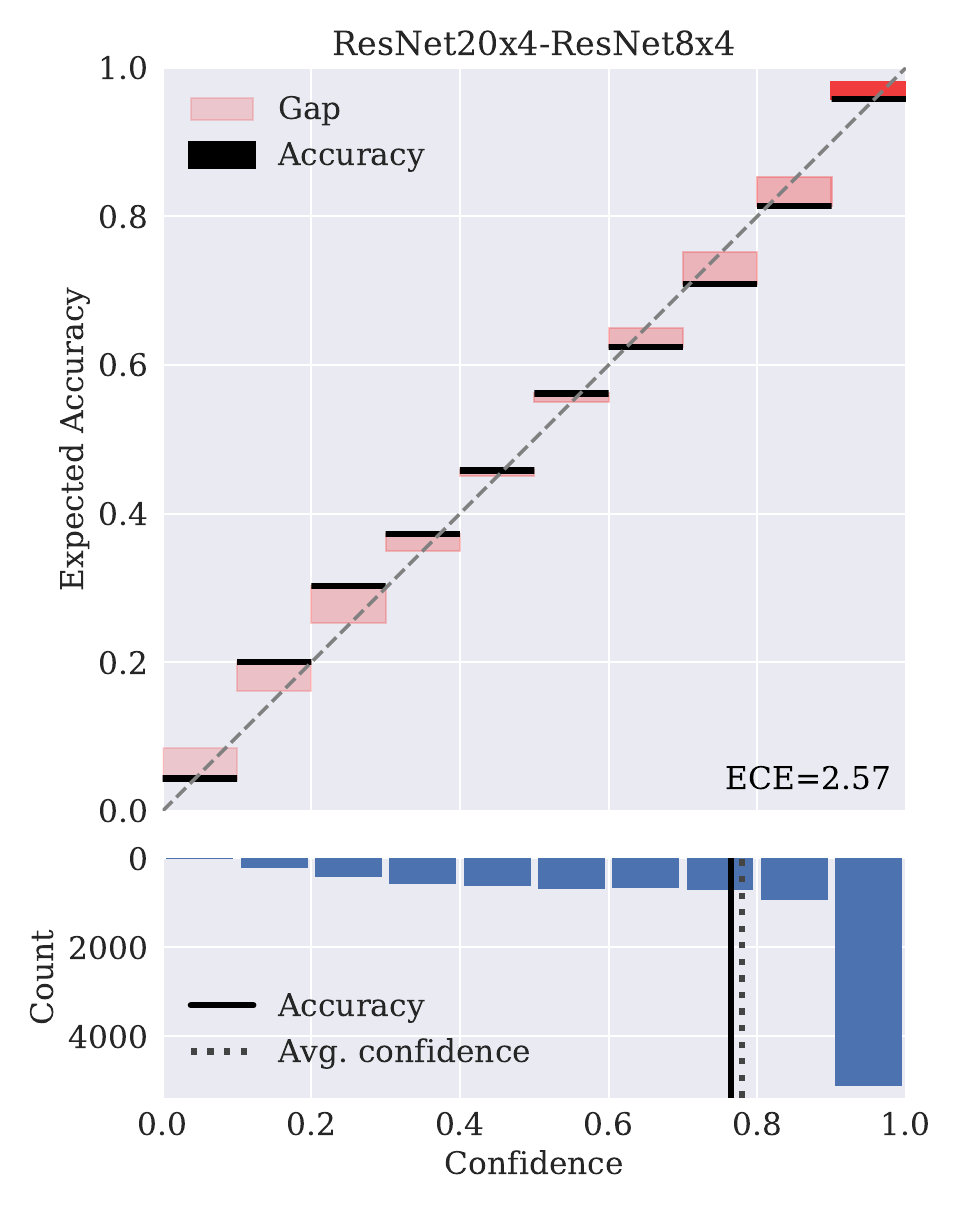}%
\label{ca3}}
\\
\subfloat[ResNet32x4]{\includegraphics[scale=0.28, trim=8 10 7 8, clip]{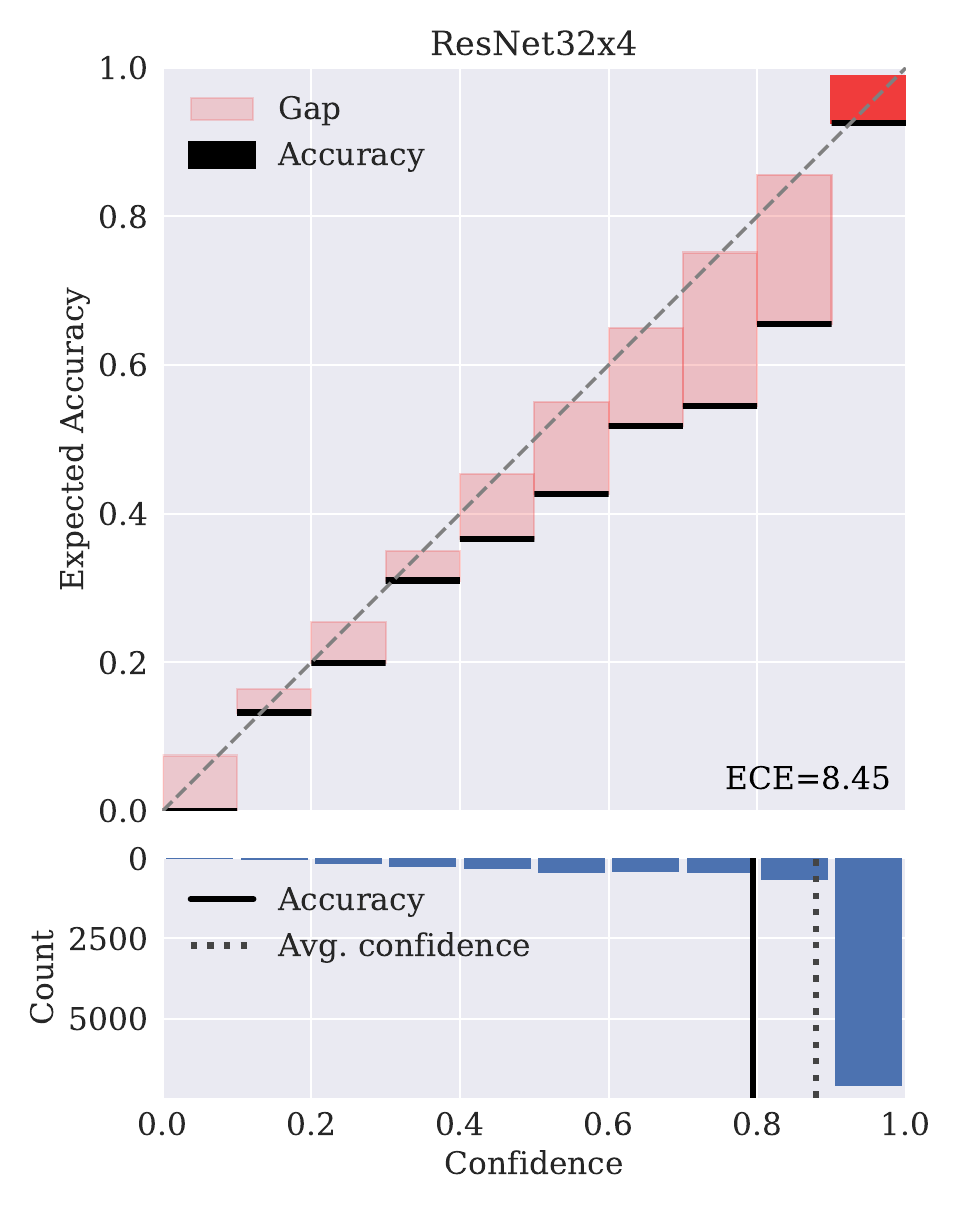}%
\label{ca4}}
\hspace{1mm}
\subfloat[ResNet8x4]{\includegraphics[scale=0.28, trim=8 10 7 8, clip]{./figures/ece_ResNet8x4.pdf}%
\label{ca5}}
\hspace{1mm}
\subfloat[w/o Projector]{\includegraphics[scale=0.28, trim=8 10 7 8, clip]{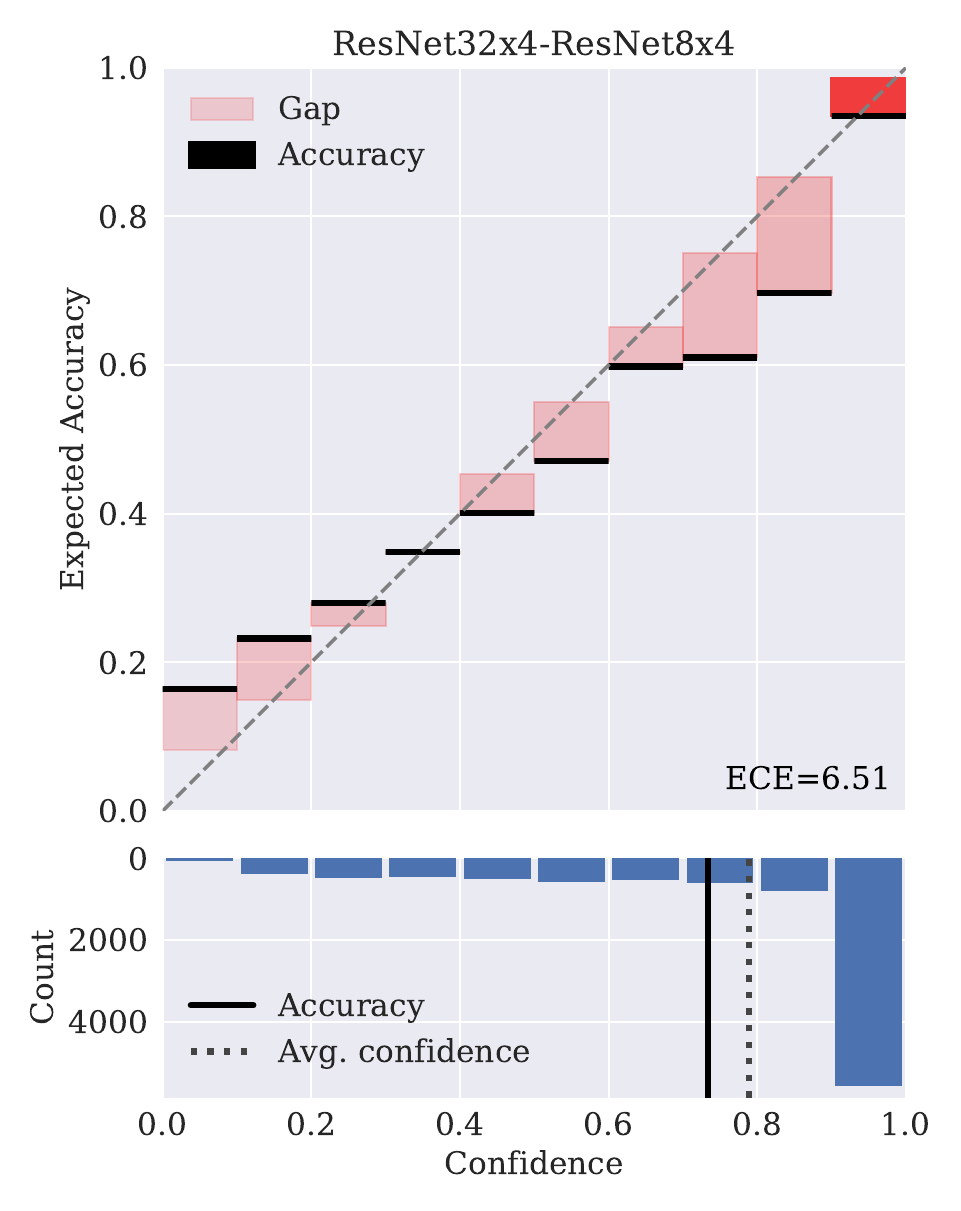}%
\label{ca6}}
\hspace{1mm}
\subfloat[w/1 Projector]{\includegraphics[scale=0.28, trim=8 10 7 8, clip]{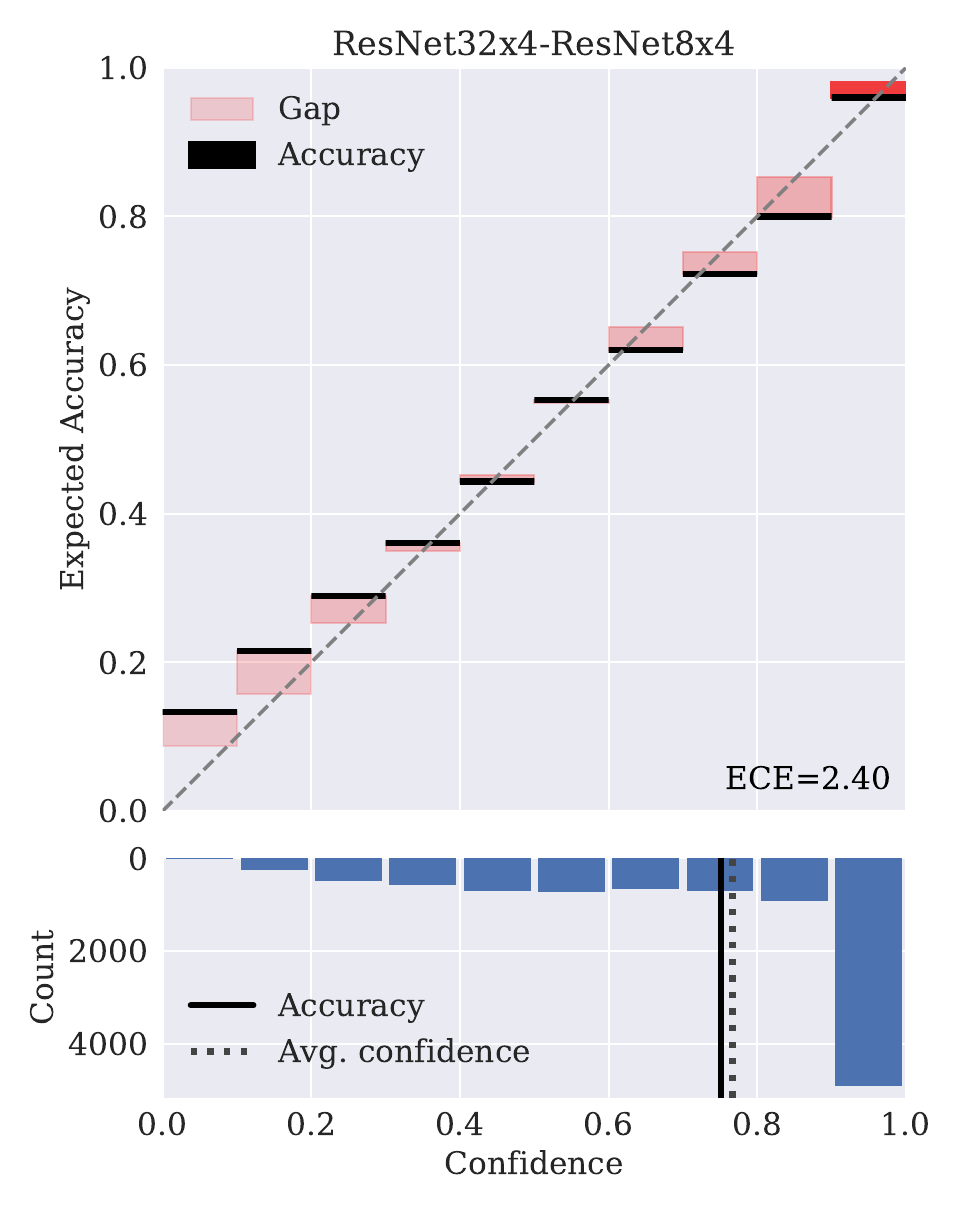}%
\label{ca7}}
\caption{Model calibration of teachers and students on CIFAR-100 testing set. The left figures (a)/(b)/(e)/(f) show the ECE of models without distillation and the right figures (c)/(d)/(g)/(h) display the ECE of models with feature distillation. }
\label{calibration}
\end{figure*}

\subsection{Projectors Increase Teacher-Student Similarity Measured by CKA}
Through the experiments with feature distillation and logit distillation, we demonstrate that adding a projector to the student is an effective way to improve classification performance. The rationale behind the use of projects is to refine the teacher's information to a level that suits the intrinsic ability of the student. In addition to the investigation of training and testing accuracies, we use the CKA similarity for measurement to further verify whether the projector can effectively extract the teacher's essential knowledge and convey it to the student. CKA similarity has been widely used to measure the similarities of features derived from different layers of a network or generated by different network architectures \cite{CKA,CKA2}. With this property, it can be used to compare the similarities between the teacher and the student, measured by applying CKA on their features \cite{contrasive3,spkd4}. We compute CKA similarities as described in \cite{CKA}.
In this experiment, we monitor the Radial Basis Function (RBF) CKA similarities between student features and teacher features by using feature distillation. The results are provided in Figure \ref{ckasimilarity}. We observe that by using the projector for feature extraction, the student can effectively exploit the latent data structure hidden in the teacher's feature space according to the ascending CKA similarities. On the other hand, we find that the student distilled with a projector has higher CKA similarities compared to the student distilled without a projector in two different teacher-student pairs. This indicates that the use of projectors results in more appropriate teacher knowledge being passed to the student, as CKA measures intrinsic notions of similarity between neural networks that transcend numeric resemblance.

\begin{figure}[t]
\centering
\subfloat[ResNet20x4-ResNet8x4]{\includegraphics[scale=0.31]{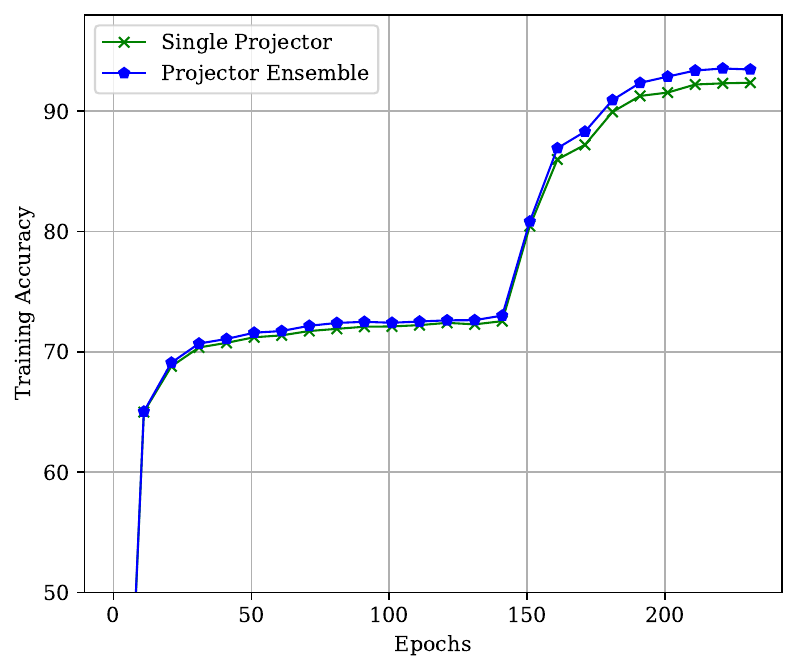}%
}
\hspace{3mm}
\subfloat[ResNet20x4-ResNet8x4]{\includegraphics[scale=0.31]{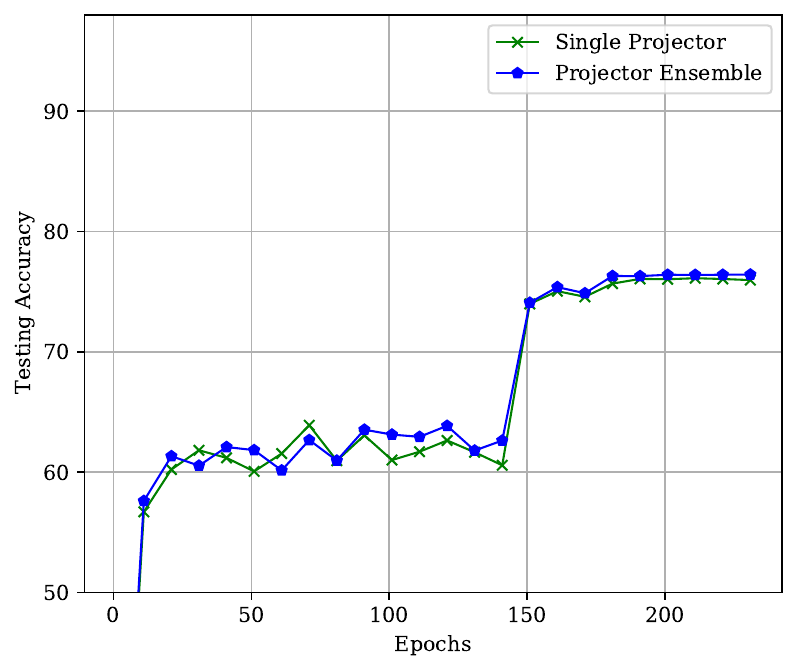}%
}
\\
\subfloat[ResNet32x4-ResNet8x4]{\includegraphics[scale=0.31]{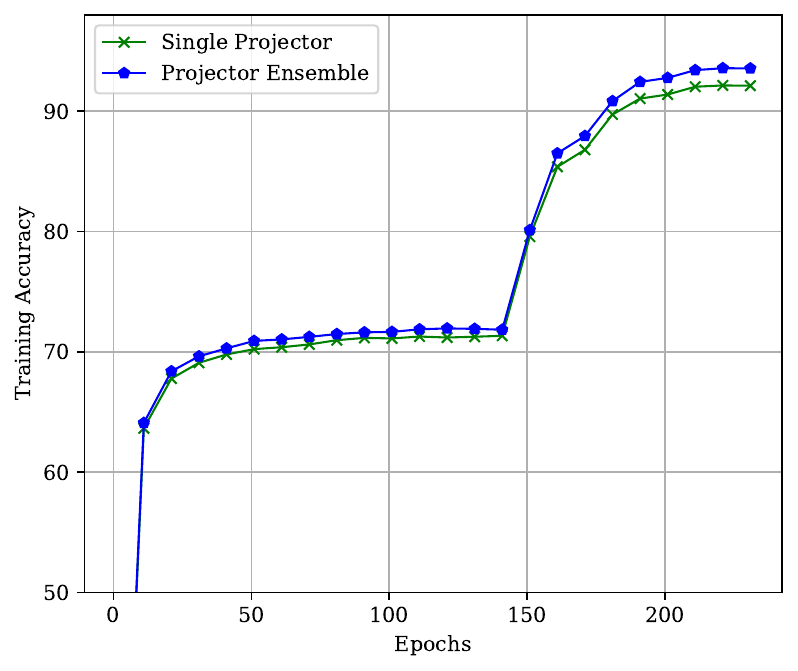}%
}
\hspace{3mm}
\subfloat[ResNet32x4-ResNet8x4]{\includegraphics[scale=0.31]{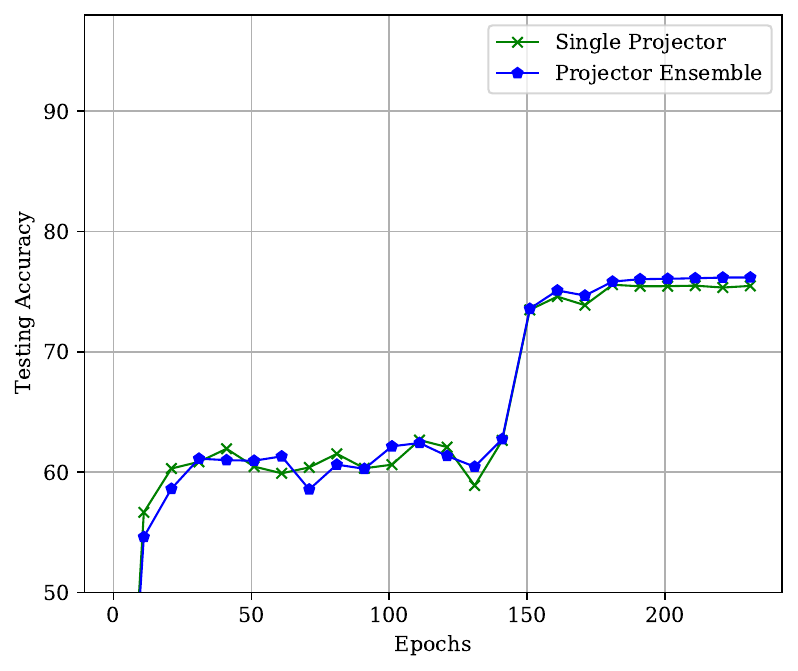}%
}
\caption{Training and testing accuracy curves by using feature distillation on CIFAR-100 dataset with different teacher-student pairs.}
\label{3projtraining}
\end{figure}
\subsection{Projectors Mitigate Teachers' Overconfidence}
Recent work indicates that modern DNNs fail to align the accuracy with the corresponding prediction confidence \cite{calibration}. Model calibration is closely related to safety in the areas such as medical analysis \cite{calibration2} and autonomous driving \cite{calibration1}. In these applications, deep models are required to obtain accuracies that are consistent with their prediction confidences. In addition to ensuring safety in certain areas, model calibration is also related to the generalization ability of networks as shown in recent research \cite{calibration3,pami3}. In this section, we investigate the calibration property of the teacher and the student. As mentioned in the existing work \cite{calibration}, increasing the size of the model tends to degrade the model's calibration. On the other hand, since the student learns representations under the guidance of the teacher, it is interesting to see whether the poor calibration ability of the teacher will be transferred to the student. To this end, we first evenly divide the prediction scores into $l$ bins and compute the Expected Calibration Error (ECE) as follows:
\begin{equation}
    \mathcal{M}_{ECE} = \sum_{i=1}^{l}\frac{|B_i|}{n_{test}}|\mathcal{M}_{ACC}(B_i)-\mathcal{M}_{CON}(B_i)|,
    \label{ece}
\end{equation}
where $n_{test}$ is the number of testing samples, and $B_i$ is the index set of testing samples that have maximum prediction scores in the $i$-th bin. In addition,
\begin{equation}
    \mathcal{M}_{ACC}(B_i) = \frac{1}{|B_i|}\sum_{j\in B_i}\textbf{1}(\hat{y}^j=y^j),
    \label{acc}
\end{equation}
where $\hat{y}^j$ is the predicted class and $y^j$ is the ground-truth class of the $j$-th sample, respectively.
\begin{equation}
   \mathcal{M}_{CON}(B_i) = \frac{1}{|B_i|}\sum_{j\in B_i}h^j,
    \label{con}
\end{equation}
where $h^j$ is the highest prediction score of the $j$-th sample. ECE is used to measure the gap between the prediction confidence and the corresponding accuracy. Therefore, the lower the ECE value, the better the model's calibration. We compute the ECE of teachers and students and depict the reliability diagram\footnote{https://github.com/hollance/reliability-diagrams/tree/master} in Figure \ref{calibration}. From the left figures, we find that larger models tend to be over-confident (the expected accuracy is higher than the actual accuracy) and have higher ECE compared to smaller models. According to the results in the right figures, it is shown that the over-confident property of the teacher is slightly mitigated by the student distilled without a projector. Surprisingly, the student with a projector significantly reduces ECE. We conjecture that this is because the student with a projector does not need to exactly mimic the teacher's representations in the original feature space and the projector refines the teacher's information with better confidence calibration.

\begin{figure}[!t]
\centering
\subfloat[ResNet20x4-ResNet8x4]{\includegraphics[scale=0.31]{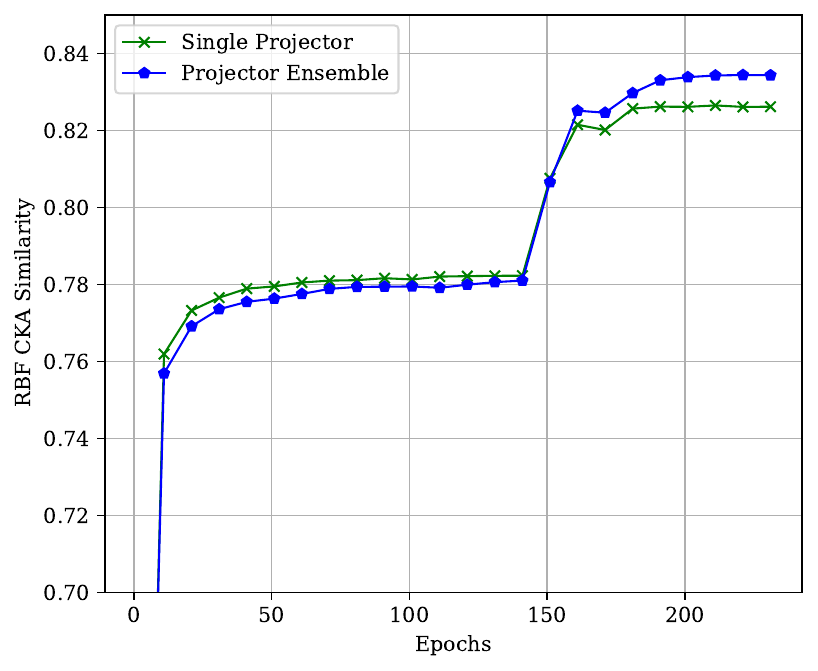}%
\label{cka13proj}}
\hspace{2mm}
\subfloat[ResNet32x4-ResNet8x4]{\includegraphics[scale=0.31]{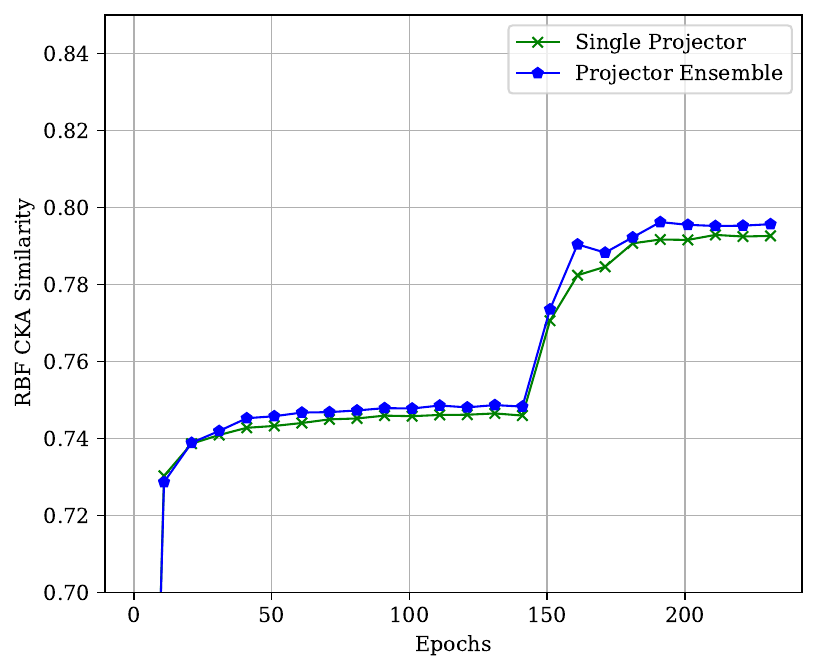}%
\label{cka33proj}}
\caption{CKA similarities between student features $S$ and teacher features $T$ on CIFAR-100 training set by distillation with a projector and with projector ensemble.}
\label{ckasimilarity3proj}
\end{figure}

\begin{algorithm}[t]
\caption{Feature Distillation via Projector Ensemble.} \label{alg:algorithm}
\textbf{Input:} The pre-trained teacher, the structure of the student, training data $X$, and labels. \\
\textbf{Parameter:} Total iterations $N$, hyper-parameter $\alpha$ and the number of projectors $q$.\\
\textbf{Initialization:} Initialize different projectors and the student.
\\
\textbf{Training:} 
\begin{algorithmic}[1] 
\FOR{$i=1\rightarrow N$}
\STATE Sample a mini-batch data from $X$.
\STATE Generate $S$, $T$, and the student's prediction by forward propagation.
\STATE Update projectors and the student network by minimizing objective (\ref{totaloss2}).
\ENDFOR
\end{algorithmic}
\textbf{Output}: The distilled student. 
\label{algorithm1}
\end{algorithm}
\subsection{Ensemble: A Better Design for Projectors in Feature Distillation}
The above analysis suggests that projectors can boost the distillation performance of the student by refining the teacher's information. Furthermore, we discuss how to design the architecture of a projector for further improvement. There are three main ways to modify the projector (i.e. depth \cite{resnet}, width \cite{wrn}, and ensemble \cite{resnext,moe}). We empirically find that the ensemble of multiple one-layer projectors can effectively improve the feature distillation performance. We conjecture that projectors with different initialization provide different transformed features, which can capture the teacher's information from different views and are beneficial to the generalizability of the student. For instance, we show the diversity in projector expertise from the perspective of categorical sensitivity. 

\begin{figure*}[t]
\centering
\subfloat[Single Projector]{\includegraphics[scale=0.28, trim=8 10 7 8, clip]{./figures/ece_ResNet20x4-ResNet8x4_1proj.pdf}%
}
\hspace{1mm}
\subfloat[Projector Ensemble]{\includegraphics[scale=0.28, trim=8 10 7 8, clip]{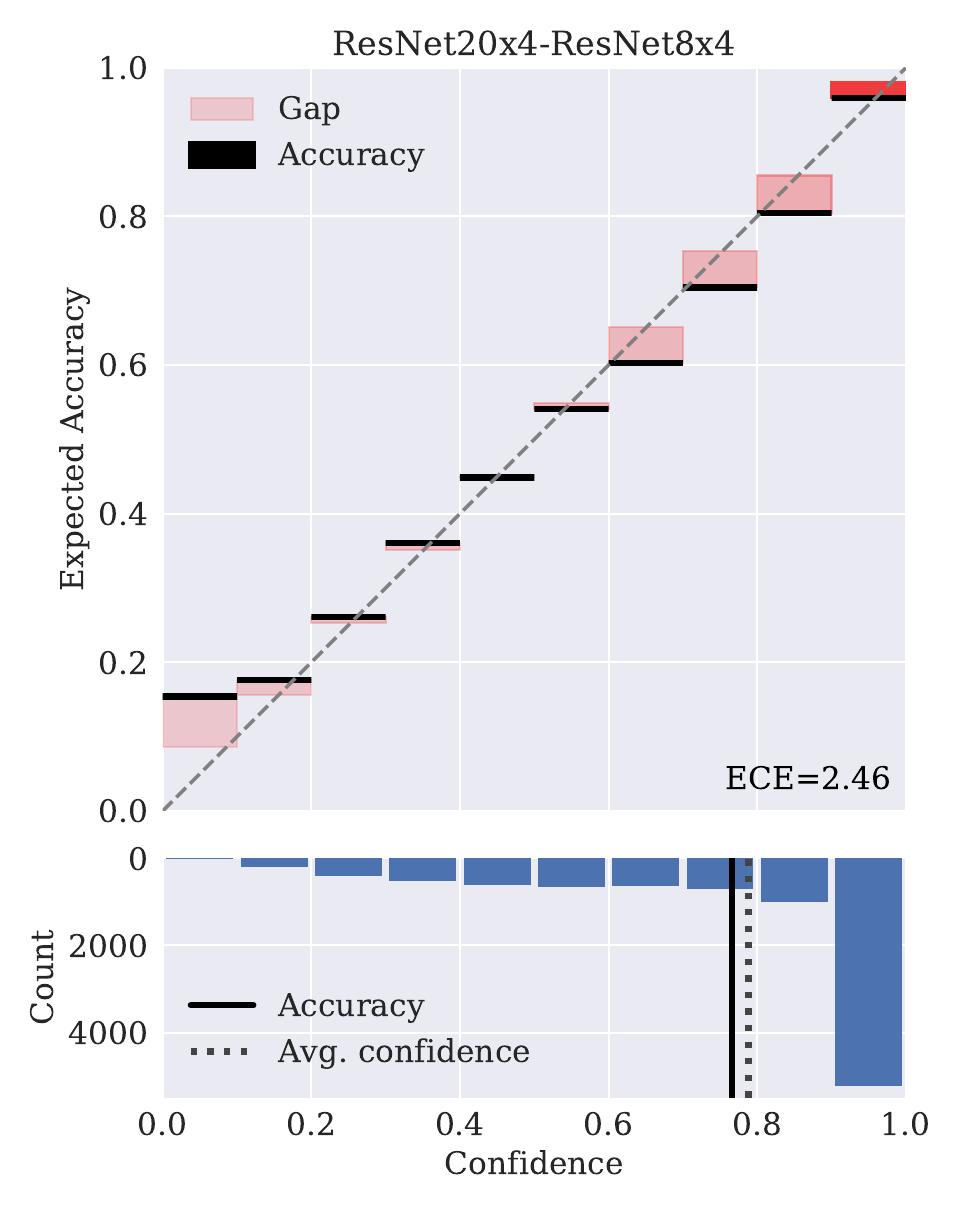}%
}
\hspace{1mm}
\subfloat[Single Projector]{\includegraphics[scale=0.28, trim=8 10 7 8, clip]{./figures/ece_ResNet32x4-ResNet8x4_1proj.pdf}%
}
\hspace{1mm}
\subfloat[Projector Ensemble]{\includegraphics[scale=0.28, trim=8 10 7 8, clip]{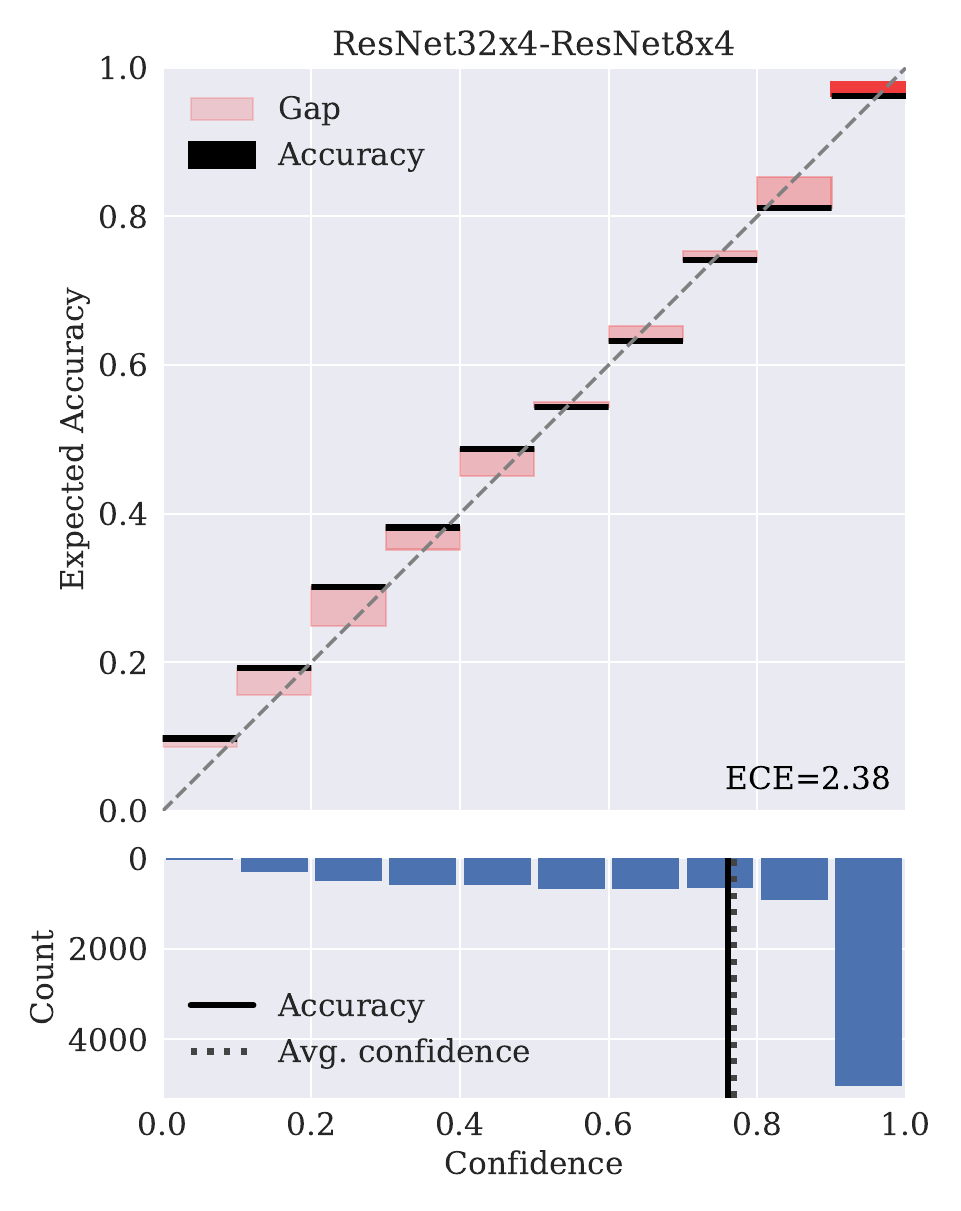}%
}
\caption{Model calibration of students trained with a projector and with projector ensemble by using feature distillation on CIFAR-100 testing set.}
\label{ece2}
\end{figure*}

\begin{table}[H]
    \caption{Top-2 class labels with the highest cosine similarities between teacher and the transformed student features.}
    \label{diversityofprojs}
    \centering
    \begin{tabular}{c|c|c|c} 
		\hline
		  	&80 Epochs	&160 Epochs	&240 Epochs \\
		\hline 
		Projector-1 	&(bear,beaver)	&(beaver,otter)	&(beaver,otter)\\
		Projector-2 	&(lobster,woman)	&(lobster,clock)	&(clock,lobster)\\
		\hline
    \end{tabular}
\end{table}
Table \ref{diversityofprojs} reports the top-2 categories that have the highest average cosine similarities between the teacher and the corresponding transformed student features, indicating that the projectors with different initialization each specialize in projecting certain categories in the image dataset. In this experiment, we use a teacher-student pair ResNet32x4-ResNet8x4 on CIFAR-100 for demonstration. Table \ref{diversityofprojs} shows that student features transformed by the first projector tend to fit teacher features with class labels `otter' and `beaver' while student features transformed by the second projector obtain higher cosine similarities with teacher features from classes `clock' and `lobster'. Therefore, it is reasonable to integrate multiple projectors to improve performance. By introducing multiple projectors, the Modified Direction Alignment (MDA) loss is as follows:
\begin{equation}
    \mathcal{L}_{MDA} = 1-\frac{1}{b}\sum_{i=1}^b\frac{\langle f(s_i),t_i\rangle}{||f(s_i)||_2||t_i||_2},
    \label{mda}
\end{equation}
where $f(s_i)=\frac{1}{q}\sum_{k=1}^q g_k(s_i)$, $q$ is the number of projectors and $g_k(\cdot)$ indicates the $k$-th projector. The overall objective function is as follows:
\begin{equation}
    \mathcal{L}_{total} = \mathcal{L}_{CE} + \alpha \mathcal{L}_{MDA}.
    \label{totaloss2}
\end{equation}
The details on performing distillation with the proposed method are shown in Algorithm \ref{algorithm1}. Similarly, we investigate the advantages of using a projector ensemble from the perspectives of the trade-off between training and testing accuracies, CKA similarities, and ECE values. In these experiments, we distill students with a projector ensemble by setting $q=3$. Figure \ref{3projtraining} demonstrates that the training and testing accuracies are improved by integrating multiple projectors to refine the teacher's information. In addition, Figures \ref{ckasimilarity3proj} and \ref{ece2} illustrate that the proposed projector ensemble method further increases the CKA similarities and reduces the ECE.

\section{Experiments} \label{experimentsection}
We conduct comprehensive experiments to evaluate the performance of different methods and the effectiveness of the proposed projector ensemble-based feature distillation, on the image classification tasks. Implementation details are as follows:

\textbf{Baselines. }We select representative distillation methods in various categories for comparisons, including logit-based method KD \cite{kd}, similarity-based methods CC \cite{cc}, SP \cite{sp}, and Relational Knowledge Distillation (RKD) \cite{rkd}, feature-based methods FitNets \cite{fitnets}, FT \cite{ft}, CRD \cite{crd}, CID \cite{cid} and SRRL \cite{srrl2,srrl}. The logit-based and similarity-based methods are projector-free and the feature-based methods require additional projectors. FitNets and SRRL use convolutional kernels to transform the student features. FT adopts an auto-encoder to extract the latent feature representations of the student and the teacher. CRD maps student and teacher features into a low-dimensional space while CID maps student features into the teacher space with linear projections. For simplicity, the proposed method constructs the projector by combining a linear projection and the ReLU function. In the reported results, Ours indicates the proposed projector ensemble-based feature distillation method. In addition, five recently proposed methods, namely Attention-based Feature Distillation (AFD) \cite{afd}, Knowledge Review (KR) \cite{kr}, DKD \cite{kd6}, Masked Generative Distillation (MGD) \cite{mgd} and Inverse Probability Weighting Distillation (IPWD) \cite{ipwd} are also introduced for comparisons. The results of these methods are cited from the original papers. 

\textbf{Datasets. }Two benchmark datasets are used for evaluation in our experiments. ImageNet \cite{imagenet} contains approximately 1.28 million training images and 50,000 validation images from 1,000 classes. The validation images are used for testing. Each image is resized to 224x224. CIFAR-100 \cite{cifar100} dataset includes 50,000 training images and 10,000 testing images from 100 classes. Each image is resized to 32x32. On ImageNet and CIFAR-100, we adopt the commonly used random crop and horizontal flip techniques for data augmentation. 

\textbf{Teacher-student pairs. }To validate the generalizability of different distillation methods, we select a group of popular network architectures to form different teacher-student pairs. The teacher networks include ResNet34 \cite{resnet}, DenseNet201 \cite{densenet}, VGG13 \cite{vgg}, ResNet20x4 \cite{resnet}, ResNet32x4 \cite{resnet} and ResNet50 \cite{resnet}. The student networks comprise of ResNet18 \cite{resnet}, MobileNet \cite{mobilenets}, VGG8 \cite{vgg}, ResNet8x4 \cite{resnet} and MobileNetV2 \cite{mobilenetsv2}. By combining different teacher and student networks, we can perform distillation between similar architectures (e.g. ResNet34-ResNet18) and different architectures (e.g. DenseNet201-ResNet18).

\textbf{Training. }Following the settings of previous methods\footnote{https://github.com/HobbitLong/RepDistiller}, the batch size, epochs, learning rate decay rate, and weight decay rate are 256/64, 100/240, 0.1/0.1, and 0.0001/0.0005, respectively on ImageNet/CIFAR-100. The initial learning rate is 0.1 on ImageNet, 0.01 for MobileNetV2 on CIFAR-100, and 0.05 for the other students on CIFAR-100. Besides, the learning rate drops every 30 epochs on ImageNet and drops at 150, 180, and 210 epochs on CIFAR-100. The optimizer is Stochastic Gradient Descent (SGD) with a momentum of 0.9. All the experiments are performed on an NVIDIA V100 GPU. 

\textbf{Hyper-parameters.} By following the conventions in CRD \cite{crd}, we use the same settings for the hyper-parameters of KD, CC, SP, RKD, FitNets, FT, and CRD. For SRRL and CID, the settings of hyper-parameters are provided by the corresponding authors. For the proposed projector ensemble-based feature distillation method, we set $\alpha=25$ and $q=3$ by tuning with the teacher-student pair ResNet34-ResNet18 on ImageNet. For a fair comparison, the hyper-parameters of different methods are fixed in all teacher-student pairs.

\subsection{Ablation Studies} \label{ablationsection}
This section studies the effectiveness of the proposed projector ensemble method, and how different ensemble strategies affect the performance. In this experiment, two different network architectures, i.e., VGG-style and ResNet-style networks are used for illustration in Tables \ref{ablationE}, \ref{ablationC}, \ref{widerproj}, \ref{activation} and \ref{initialization}.

\begin{table*}
    \caption{Top-1 accuracy$\pm$standard deviation ($\%$) on CIFAR-100 using horizontal ensembles with different teacher-student pairs. 1-Proj, 2-Proj, 3-Proj, and 4-Proj indicate the number of single-layer projectors in the ensemble.}
    \label{ablationE}
    \centering
    \begin{tabular}{c|cccccc|c}
    \hline
    Pair &Student &w/o Proj &1-Proj &2-Proj &3-Proj &4-Proj  &Teacher\\
    \hline 
    \multirow{1}{*}{VGG13-VGG8} &70.74$\pm$0.31 &73.76$\pm$0.21  &73.84$\pm$0.25  &74.21$\pm$0.31  &\textbf{74.35$\pm$0.12} &74.18$\pm$0.04  &74.64  \\
    \hline
    \multirow{1}{*}{ResNet32x4-ResNet8x4}  &72.93$\pm$0.28 &73.66$\pm$0.15  &75.14$\pm$0.21  &75.66$\pm$0.15  &\textbf{76.08$\pm$0.33}  &75.93$\pm$0.09 &79.42   \\
    \hline
    \end{tabular}
\end{table*}

\begin{table*}
    \caption{Top-1 accuracy$\pm$standard deviation ($\%$) on CIFAR-100 using deep projectors with different teacher-student pairs. 2L-MLP, 3L-MLP, and 4L-MLP represent projectors with different number of layers.}
    \label{ablationC}
    \centering
    \begin{tabular}{c|cccccc|c}
    \hline
    Pair &Student &w/o Proj &1-Proj &2L-MLP &3L-MLP &4L-MLP &Teacher \\
    \hline 
    \multirow{1}{*}{VGG13-VGG8} &70.74$\pm$0.31  &73.76$\pm$0.21  &\textbf{73.84$\pm$0.25}  &73.31$\pm$0.26  &73.02$\pm$0.34 &72.73$\pm$0.09  &74.64   \\
    \hline
    \multirow{1}{*}{ResNet32x4-ResNet8x4}  &72.93$\pm$0.28 &73.66$\pm$0.15  &\textbf{75.14$\pm$0.21}  &75.12$\pm$0.20  &74.56$\pm$0.06 &74.30$\pm$0.21 &79.42   \\
    \hline
    \end{tabular}
\end{table*}

\begin{table}
    \caption{Top-1 accuracy$\pm$standard deviation ($\%$)  on CIFAR-100 using deep projectors. 2Lx2-MLP and 2Lx3-MLP indicate two-layer MLPs with wider hidden layers.}
    \label{widerproj}
    \centering
    \begin{tabular}{c|c|c} 
		\hline
		Teacher &VGG13 &{ResNet32x4} \\
		Student &VGG8 &{ResNet8x4} \\
		\hline 
		2L-MLP &73.31$\pm$0.26   &75.12$\pm$0.20\\
		2Lx2-MLP &73.31$\pm$0.21   &75.00$\pm$0.25\\
		2Lx3-MLP &\textbf{73.37$\pm$0.10}  &\textbf{75.13$\pm$0.19}\\
		\hline
    \end{tabular}
\end{table}

\begin{table}
    \caption{Top-1 accuracy$\pm$standard deviation ($\%$) of our method with different activation functions on CIFAR-100.}
    \label{activation}
    \centering
    \begin{tabular}{c|c|c} 
		\hline
		Teacher &VGG13 &{ResNet32x4} \\
		Student &VGG8 &{ResNet8x4} \\
		\hline 
		w/ ReLU (Ours) &74.35$\pm$0.12   &76.08$\pm$0.33\\
		w/ GELU &\textbf{74.39$\pm$0.18}   &\textbf{76.32$\pm$0.27}\\
		w/o activation &73.46$\pm$0.42  &75.04$\pm$0.37\\
		\hline
    \end{tabular}
\end{table}

\begin{table}[t]
    \caption{Top-1 accuracy$\pm$standard deviation ($\%$) of our method with different projector initialization strategies on CIFAR-100.}
    \label{initialization}
    \centering
    \begin{tabular}{c|c|c} 
		\hline
		Teacher &VGG13 &{ResNet32x4} \\
		Student &VGG8 &{ResNet8x4} \\
		\hline 
		Default initialization (Ours) &\textbf{74.35$\pm$0.12}   &76.08$\pm$0.33\\
		He initialization &74.32$\pm$0.24   &75.78$\pm$0.16\\
		Orthogonal initialization &73.89$\pm$0.11 &76.12$\pm$0.26\\
		Mixed initialization &74.00$\pm$0.16  &\textbf{76.27$\pm$0.20}\\
		\hline
    \end{tabular}
\end{table}
\textbf{Horizontal ensemble of projectors. }Table \ref{ablationE} shows the top-1 classification accuracy of the proposed projector ensemble with a different number of projectors. On CIFAR-100, we run our method three times with different seeds to obtain the average accuracy and the corresponding standard deviation. Results in Table \ref{ablationE} verify that imposing a projector improves the distillation performance when the feature dimensions of the student and the teacher are the same. A potential reason is that the projector helps to refine the teacher's information and consequently improves the quality of student features. In addition, by integrating multiple projectors for feature transformation, the proposed method further increases the classification accuracy by a clear margin with various numbers of projectors. 

\textbf{Deep cascade of projectors. }Another common way to modify the architecture is to increase the depth of the projector. Table \ref{ablationC} demonstrates the changes in distillation performance by gradually stacking non-linear projections. In this table, 2L-MLP, 3L-MLP, and 4L-MLP are multilayer perceptrons and each layer outputs $m$-dimensional features followed by a ReLU activation. For instance, the output of 2L-MLP is $g(s_i)=\sigma(W_2\sigma(W_1s_i))$, where $W_1 \in \mathbb{R}^{m\times d}$ and $W_2 \in \mathbb{R}^{m\times m}$ are weighting matrices. It is shown that simply increasing the depth of the projector does not improve the performance of the student and tends to degrade the effectiveness of the projector. We hypothesize that with the increase of depth, the teacher's features can be overfitted by the projector. In addition, we also evaluate the effects of deep projectors with wider hidden layers as shown in Table \ref{widerproj}. In this experiment, the hidden dimensions of 2Lx2-MLP and 2Lx3-MLP are $2\times m$ and $3\times m$, respectively. Experimental results demonstrate that the improvement is marginal by increasing the width of deep projectors.

\textbf{Activation of projectors. }We further investigate the influence of the activation function. Table \ref{activation} shows that the addition of the ReLU activation function has a significant positive impact on the performance of the proposed method. The reason is that the lack of an activation function and the non-linearity introduced by it limits the diversity of the projectors, as a group of linear projections can be mathematically reduced to a single linear projection through sum-pooling, which will degrade the distillation performance. Recently, Gaussian Error Linear Units (GELU) \cite{gelu} has received attention because of its effectiveness on Transformer \cite{bert,vit}. We replace ReLU with GELU in the proposed method to test the performance. It is shown that the performance of our method can be further improved by using GELU as activation. 

\textbf{Initialization of projectors. }In our experiments, we find that simply initializing different projectors with different seeds and the default initialization method of the linear layer in PyTorch is sufficient to yield good performance. Therefore, we stick to this strategy to make the proposed method as simple as possible. We also compare the distillation performance by using different initialization methods as shown in Table \ref{initialization}. Similarly, we use VGG13-VGG8 and ResNet32x4-ResNet8x4 on CIFAR-100 for evaluation in this experiment. The initialization methods include He initialization \cite{kaimingini}, orthogonal initialization \cite{xavierini}, the default initialization in PyTorch, and mixing the above three initialization methods. Experimental results show that mixing different initialization methods has a slight impact on the results and is a potential way to further improve the distillation performance. 

\begin{table}[t]
    \caption{Top-1 accuracy$\pm$standard deviation ($\%$) on CIFAR-100 with different teacher-student pairs.}
    \label{tableCIFAR}
    \centering
    \begin{tabular}{c|c|c|c|c}
    \hline
    Teacher     &VGG13  &ResNet32x4  &ResNet50 &ResNet50        \\
    Student    &VGG8    &ResNet8x4 &VGG8    &MobileNetV2      \\
    \hline
    Teacher   &74.64 &79.42 &79.34 &79.34 \\
    Student   &70.74$\pm$0.31 &72.93$\pm$0.28 &70.74$\pm$0.31 &65.03$\pm$0.09\\
    KD     &72.98$\pm$0.19 &73.33$\pm$0.25 &73.81$\pm$0.13 &67.35$\pm$0.32\\
    CC   &70.71$\pm$0.24 &72.97$\pm$0.17 &70.25$\pm$0.12 &65.43$\pm$0.15\\
    SP  &72.68$\pm$0.19 &72.94$\pm$0.23 &73.34$\pm$0.34 &68.08$\pm$0.38\\
    RKD   &71.48$\pm$0.05 &71.90$\pm$0.11 &71.50$\pm$0.07 &64.43$\pm$0.42  \\
    FitNets   &71.02$\pm$0.31 &73.50$\pm$0.28 &70.69$\pm$0.22 &63.16$\pm$0.47\\
    FT  &70.58$\pm$0.08 &72.86$\pm$0.12 &70.29$\pm$0.19 &60.99$\pm$0.37 \\
    CRD    &73.94$\pm$0.22 &75.51$\pm$0.18 &74.30$\pm$0.14 &69.11$\pm$0.28\\
    SRRL  &73.44$\pm$0.07 &75.33$\pm$0.04 &74.23$\pm$0.08 &68.41$\pm$0.54\\
    Ours  &\textbf{74.35$\pm$0.12} &\textbf{76.08$\pm$0.33} &\textbf{74.58$\pm$0.22} &\textbf{69.81$\pm$0.42} \\
    \hline
    \end{tabular}
\end{table}

\subsection{Results on CIFAR-100} \label{cifarsection}
Table \ref{tableCIFAR} reports the experimental results on CIFAR-100 with different teacher-student pairs. Since CID requires different hyper-parameters for different pairs to achieve good performance, we omit it for comparisons on CIFAR-100. Among the projector-free distillation methods, the logit-based method KD shows better performance compared to the similarity-based methods CC, SP and RKD. Furthermore, KD outperforms the projector-based methods FitNets and FT in most cases. Since FitNets is designed to distill the intermediate features, its performance is unstable for teacher-student pairs using different architectures. FT uses an auto-encoder as the projector to extract latent representations of the teacher, which may disturb the discriminative information to some extent and consequently degrade the performance. The recently proposed feature distillation methods CRD and SRRL show competitive performance compared to the previous methods by distilling the last layer of features. By harnessing the power of distilling the last features and projector ensemble, the proposed method achieves the highest top-1 accuracy in four teacher-student pairs and consistently obtains competitive performance. 

\subsection{Results on ImageNet}
The classification performance of the students distilled by different methods on ImageNet is listed in Tables \ref{tableImageNet} and \ref{tableImageNet2}. Compared to the experimental settings in previous methods \cite{crd,srrl,cid}, we introduce more teacher-student pairs for evaluation on this large-scale dataset so that the generalizability of different methods can be better evaluated. As presented in the tables, feature distillation methods (CRD, SRRL, CID, and our method) outperform both the logit-based method (KD) and the similarity-based method (SP) in most teacher-student pairs.

\begin{figure}[t]
    \centering
	\subfloat[ResNet34-ResNet18]{\includegraphics[scale=0.36]{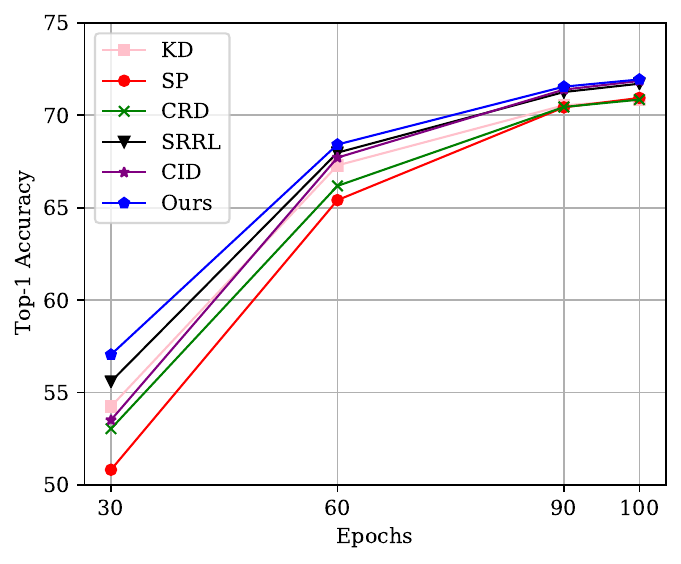}}
        \hspace{3mm}
	\subfloat[ResNet50-MobileNet]{\includegraphics[scale=0.36]{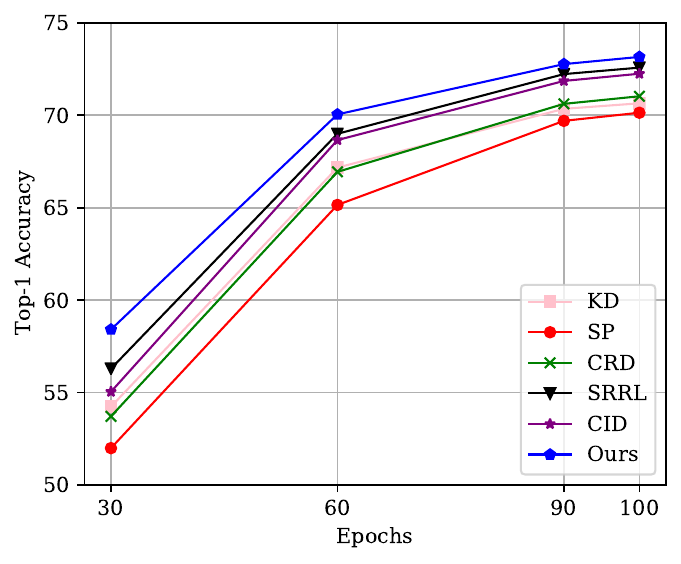}}
	\caption{Top-1 accuracy of different methods on ImageNet at different epochs with different teacher-student pairs.}
	\label{converge}
\end{figure}

One major difference between CRD and the other feature distillation methods is the way of feature transformation. CRD transforms teacher and student features simultaneously while the other methods only transform student features. By solely mapping student features into the teacher's space, the original teacher feature distribution can be preserved without losing discriminative information. Therefore, SRRL, CID, and our method obtain better performance than CRD. Besides, the proposed projector ensemble method consistently outperforms the state-of-the-art methods SRRL and CID with different teacher-student pairs. With pair DenseNet201-MobileNet, the proposed method obtains 0.96$\%$ and 0.50$\%$ improvements compared to the second-best method in terms of top-1 and top-5 accuracy, respectively.  MobileNet (4.2M parameters) distilled by our method can obtain similar performance and reduce about 80$\%$ of the parameters compared to the ResNet34 (21.8M parameters). Figure \ref{converge} reports the top-1 accuracy of different methods with different training epochs. It is shown that the proposed method converges faster than the other distillation methods.

\begin{table}[t]
    \caption{Top-1 and Top-5 classification accuracy ($\%$) of different methods on ImageNet with different teacher-student pairs.}
    \label{tableImageNet}
    \centering
    \begin{tabular}{c|cc|cc}
    \hline
    Teacher     &\multicolumn{2}{c|}{ResNet34}    &\multicolumn{2}{c}{ResNet50}        \\
    Student    &\multicolumn{2}{c|}{ResNet18}     &\multicolumn{2}{c}{MobileNet}       \\
    \hline
    Accuracy &Top-1 ($\%$) &Top-5 ($\%$) &Top-1 ($\%$) &Top-5 ($\%$) \\
    \hline
    Teacher &73.31 &91.41 &76.13 &92.86 \\
    Student &69.75 &89.07 &69.06 &88.84   \\
    KD     &70.83 &90.15 &70.65 &90.26 \\
    SP  &70.94 &89.83 &70.14 &89.64 \\
    CRD    &70.85 &90.12 &71.03 &90.16 \\
    CID  &71.86 &90.63 &72.25 &90.98 \\
    SRRL  &71.71 &90.58 &72.58 &91.05 \\
    Ours &\textbf{71.94} &\textbf{90.68} &\textbf{73.16} &\textbf{91.24} \\
    \hline
    \end{tabular}
\end{table}

\begin{table}[t]
    \caption{Top-1 and Top-5 classification accuracy ($\%$) of different methods on ImageNet with different teacher-student pairs.}
    \label{tableImageNet2}
    \centering
    \begin{tabular}{c|cc|cc}
    \hline
    Teacher     &\multicolumn{2}{c|}{DenseNet201}    &\multicolumn{2}{c}{DenseNet201}        \\
    Student    &\multicolumn{2}{c|}{ResNet18}     &\multicolumn{2}{c}{MobileNet}       \\
    \hline
    Accuracy &Top-1 ($\%$) &Top-5 ($\%$) &Top-1 ($\%$) &Top-5 ($\%$) \\
    \hline
    Teacher   &76.89 &93.37 &76.89 &93.37  \\
    Student &69.75 &89.07 &69.06 &88.84   \\
    KD     &70.38 &90.12 &69.98 &89.93 \\
    SP  &70.75 &90.01 &70.34 &89.63 \\
    CRD    &70.87 &89.86 &70.82 &90.09 \\
    CID  &71.99 &90.64 &71.90 &90.97 \\
    SRRL  &71.76 &90.80 &72.28 &90.90 \\
    Ours &\textbf{72.29} &\textbf{90.99} &\textbf{73.24} &\textbf{91.47} \\
    \hline
    \end{tabular}
\end{table}

In \cite{takd}, the authors observe that a better teacher may fail to distill a better student. Such phenomenon also exists by comparing the results in Tables \ref{tableImageNet} and \ref{tableImageNet2}. For example, compared to the pair ResNet50-MobileNet, most of the methods distill a worse student by using a better network DenseNet201 as the teacher. One plausible explanation for this phenomenon is that the knowledge of a better teacher is more complex and more difficult to learn. To alleviate this problem, TAKD \cite{takd} introduces some smaller assistant networks to facilitate training. Densely Guided Knowledge Distillation (DGKD) \cite{densetakd} further extends TAKD with dense connections between different assistants. However, the training costs of these methods are greatly increased by using the assistant networks. As shown in Tables \ref{tableImageNet} and \ref{tableImageNet2}, the proposed method has the potential to alleviate this problem without introducing additional networks.

We conduct an ablation study on ImageNet and the results are reported in Table \ref{tableAblation}. In this experiment, we select two teacher-student pairs for illustration. ResNet34 and ResNet18 have similar network architectures. ResNet50 and MobileNet have different network architectures, which can be used to observe the performance of the proposed projector ensemble method when the feature dimensions of the teacher and the student are different. Results in Table \ref{tableAblation} are consistent with that in Table \ref{ablationE} from the following two aspects. Firstly, adding a projector to the student has a positive impact on feature distillation when the teacher and the student have identical feature dimensions. Secondly, integrating multiple projectors for feature transformation can further improve the performance in the cases of distillation between similar architectures, and different architectures.

Five recently proposed methods are also added for comparisons as illustrated in Table \ref{crosslayer}. Unlike methods that utilize the last layer of features for distillation (CRD, SRRL, CID, and ours), AFD and KR propose to extract information from multiple layers of features. Table \ref{crosslayer} shows that the proposed method performs better than AFD and KR with different teacher-student pairs in terms of the comparisons of top-1 and top-5 accuracy, which indicates that using the last layer of features is sufficient to obtain good distillation performance on ImageNet.

\subsection{Training Complexity}
\begin{figure}[t]
    \centering
	\subfloat{\includegraphics[scale=0.35]{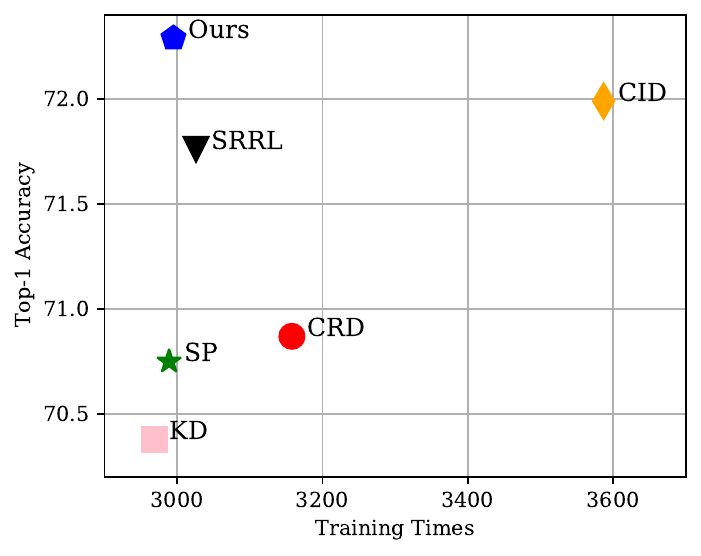}}
	\hspace{3mm}
	\subfloat{\includegraphics[scale=0.35]{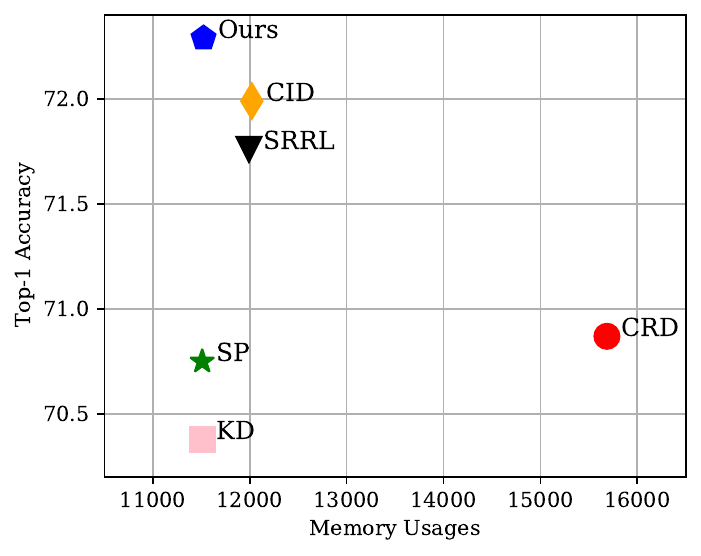}}
	\caption{Training times (in second) of one epoch and peak GPU memory usages (MB) of different methods on ImageNet with DenseNet201-ResNet18.}
	\label{tabletime}
\end{figure}

We compare the training costs of different methods on ImageNet with DenseNet201-ResNet18 in Figure \ref{tabletime} from the perspectives of training times of one epoch and peak GPU memory usage. Since KD and SP are projector-free methods, their training costs are lower than that of feature distillation methods. The training cost of the proposed projector ensemble method is slightly higher than KD and SP because we use multiple projectors to improve the optimization of the student. On the other hand, the proposed method only uses a naive direction alignment loss to distill the knowledge of the teacher. Therefore, the computation complexity and memory usages are lower compared to the other feature-based methods. To be specific, CID combines feature matching loss and interventional distillation loss \cite{causal1,causal2,causal3} in one objective, which leads to large computation complexity. CRD performs contrastive learning \cite{contrasive1,contrasive2} on feature distillation, which requires more GPU memories to save the negative samples.

\begin{table}[t]
    \caption{Ablation study on ImageNet by using horizontal ensembles with teacher-student pairs ResNet34-ResNet18 and ResNet50-MobileNet.}
    \label{tableAblation}
    \centering
    \begin{tabular}{c|cc|cc}
    \hline
    Teacher     &\multicolumn{2}{c|}{ResNet34}    &\multicolumn{2}{c}{ResNet50}        \\
    Student    &\multicolumn{2}{c|}{ResNet18}     &\multicolumn{2}{c}{MobileNet}       \\
    \hline
    Accuracy &Top-1 ($\%$) &Top-5 ($\%$) &Top-1 ($\%$) &Top-5 ($\%$) \\
    \hline    
    Teacher &73.31 &91.41 &76.13 &92.86 \\
    Student &69.75 &89.07 &69.06 &88.84   \\
    w/o Proj &71.63  &90.33 &-- &--   \\
    1-Proj &71.67  &90.59 &72.75 &91.36   \\
    2-Proj &71.82 &90.66 &73.15  &91.48   \\
    3-Proj &\textbf{71.94}  &90.68 &73.16  &91.24   \\
    4-Proj &71.83 &\textbf{90.72} &\textbf{73.29} &\textbf{91.49}   \\
    \hline
    \end{tabular}
\end{table}

\begin{table}[t]
    \caption{Comparisons of the proposed method and several recently proposed distillation methods.}
    \label{crosslayer}
    \centering
    \begin{tabular}{c|cc|cc}
    \hline
    Teacher     &\multicolumn{2}{c|}{ResNet34}    &\multicolumn{2}{c}{ResNet50}        \\
    Student    &\multicolumn{2}{c|}{ResNet18}     &\multicolumn{2}{c}{MobileNet}       \\
    \hline
    Accuracy &Top-1 ($\%$) &Top-5 ($\%$) &Top-1 ($\%$) &Top-5 ($\%$) \\
    \hline    
    Teacher &73.31 &91.41 &76.13 &92.86 \\
    Student &69.75 &89.07 &69.06 &88.84   \\
    AFD &71.38 &-- &-- &--   \\
    KR &71.61 &90.51 &72.56 &91.00   \\
    DKD &71.70 &90.41 &72.05 &91.05 \\
    MGD &71.80 &90.40 &72.59 &90.94 \\
    IPWD &71.88 &90.50 &72.65 &91.08   \\
    Ours &\textbf{71.94} &\textbf{90.68} &\textbf{73.16} &\textbf{91.24}    \\
    \hline
    \end{tabular}
\end{table}

\subsection{Transferability}
To investigate the transferability of the distilled student, we introduce two datasets for evaluation, including CUB-200-2011 \cite{cub200} and Cars-196 \cite{car196}. CUB-200-2011 comprises 5,994 training images and 5,794 testing images from 200 classes. Cars-196 contains 8,144 training images and 8,041 testing images belonging to 196 classes. In this experiment, we first train the student networks on ImageNet by using different feature distillation methods. Then, we freeze the parameters of the backbones of the distilled students and retrain the last classification layers on CUB-200-2011 and Cars-196. In this way, we can evaluate the generalizability of the students pre-trained on a large-scale dataset.

The top-5 classification accuracy of two different student networks on CUB-200-2011 and Cars-196 are reported in Table \ref{tableCUBcar}. According to the results on CUB-200-2011, the student networks distilled by different feature distillation methods generalize better on a different dataset compared to the student networks without using distillation. Although CID obtains higher top-1 accuracy than SRRL on ImageNet with DenseNet201-ResNet18, the student distilled by CID underperforms the student distilled by SRRL. A plausible reason is that the student distilled by CID overfits the teacher's outputs on ImageNet by using multiple losses. Since the proposed method achieves feature distillation with a simple direction alignment loss and an ensemble of projectors, the transferability of the distilled students is better compared to existing methods. 

On the contrary, results on Cars-196 show that existing feature distillation methods may distill a worse student from the perspective of generalization ability. With DenseNet201-ResNet18, the student networks distilled by CRD, SRRL, and CID perform worse than the student network without using distillation. Our method consistently generates a better student on Cars-196 as shown in Table \ref{tableCUBcar}. The ResNet18 distilled by the proposed projector ensemble method outperforms the vanilla ResNet18 by 3.11$\%$.

\section{Conclusion} \label{conclusection}
This paper explores the effects of projectors in knowledge distillation and accordingly proposes a projector ensemble method to improve the student's performance. We demonstrate that by adding a projector to refine the teacher's information, the quality of student features can be greatly improved from the perspectives of classification accuracy, CKA similarities and model calibration. Based on the positive effects of projectors, we propose to randomly initialize multiple projectors and achieve feature transformation in an ensemble learning manner. In this way, projectors with different initialization can capture the teacher's knowledge from different views and further improve the classification performance, similarity preservation and model calibration. Extensive experiments on benchmark datasets also illustrate the superior performance of the proposed method compared to the state-of-the-art distillation methods.

\begin{table}[t]
    \caption{Top-5 accuracy ($\%$) on CUB-200-2011 and Cars-196 with different teacher-student pairs.}
    \label{tableCUBcar}
    \centering
    \begin{tabular}{c|cc|cc}
    \hline
    Teacher     &\multicolumn{2}{c|}{ResNet50}    &\multicolumn{2}{c}{DenseNet201}        \\
    Student    &\multicolumn{2}{c|}{MobileNet}     &\multicolumn{2}{c}{ResNet18}       \\
    \hline
    Dataset &CUB-200 &Cars-196 &CUB-200 &Cars-196 \\
    \hline    
    w/o distillation &89.62 &79.33 &88.95 &76.47\\
    CRD &89.61 &79.62 &88.97 &74.77 \\
    CID &90.21 &77.59 &89.43 &73.54 \\
    SRRL &90.26 &80.41 &89.78 &76.34 \\
    Ours &\textbf{91.14} &\textbf{82.27} &\textbf{90.54} &\textbf{79.58}    \\
    \hline
    \end{tabular}
\end{table}

\ifCLASSOPTIONcaptionsoff
  \newpage
\fi

\bibliographystyle{IEEEtran}
\bibliography{references.bib}

\end{document}